CrossMark

# A constrained ℓ1 minimization approach for estimating multiple sparse Gaussian or nonparanormal graphical models

Beilun Wang[1] · Ritambhara Singh[1] · Yanjun Qi[1]

**Abstract** Identifying context-specific entity networks from aggregated data is an important task, arising often in bioinformatics and neuroimaging applications. Computationally, this task can be formulated as jointly estimating multiple different, but related, sparse undirected graphical models (UGM) from aggregated samples across several contexts. Previous joint-UGM studies have mostly focused on sparse Gaussian graphical models (sGGMs) and can't identify context-specific edge patterns directly. We, therefore, propose a novel approach, SIMULE (detecting Shared and Individual parts of MULtiple graphs Explicitly) to learn multi-UGM via a constrained ℓ1 minimization. SIMULE automatically infers both specific edge patterns that are unique to each context and shared interactions preserved among all the contexts. Through the ℓ1 constrained formulation, this problem is cast as multiple independent subtasks of linear programming that can be solved efficiently in parallel. In addition to Gaussian data, SIMULE can also handle multivariate Nonparanormal data that greatly relaxes the normality assumption that many real-world applications do not follow. We provide a novel theoretical proof showing that SIMULE achieves a consistent result at the rate $O(\log(Kp)/n_{tot})$. On multiple synthetic datasets and two biomedical datasets, SIMULE shows significant improvement over state-of-the-art multi-sGGM and single-UGM baselines (SIMULE implementation and the used datasets @ https://github.com/QData/SIMULE).

**Keywords** Graphical model · Multi-task learning · Computational biology

---

---

Beilun Wang
bw4mw@virginia.edu

Yanjun Qi
yq2h@virginia.edu

Ritambhara Singh
rs3zz@virginia.edu

[1]   University of Virginia, Charlottesville, VA, USA

Springer

## List of symbols

| | |
|---|---|
| $\Sigma$ | Covariance matrix |
| $\mu$ | Mean vector in the Gaussian distribution |
| $\Omega$ | Precision matrix |
| $\mathbf{X}$ | Data sample matrix |
| $X_j$ | A random variable follows the Gaussian Distribution |
| $\Sigma^{(i)}$ | $i$th Covariance matrix |
| $\Omega^{(i)}$ | $i$th Precision matrix in a multi-task setting |
| $\Omega_S$ | Shared pattern among all precision matrices in a multi-task setting |
| $\Omega_I^{(i)}$ | Individual part of $i$th Precision matrix in a multi-task setting |
| $\mathbf{X}^{(i)}$ | $i$th data sample matrix in a multi-task setting |
| $n_i$ | Number of samples in $i$th data matrix |
| $n_{tot}$ | Total number of samples in a multi-task setting |
| $\beta$ | A column of $\Omega$ |
| $\beta^{(i)}$ | A column of $\Omega^{(i)}$ |
| $\mathbf{x}$ | A $p$-dimensional sample |
| $\mathbf{e}_j$ | $(0, \ldots, 1, \ldots, 0)^T$ |
| $\theta$ | A parameter used in linear programming formulation [Eq. 10], $[\beta^{(1)^T}, \ldots, \beta^{(i)^T}, \ldots, \beta^{(K)^T}, \varepsilon K(\beta^s)^T]^T$ |
| $\mathbf{A}^{(i)}$ | A parameter used in linear programming formulation [Eq. 10], $[0, \ldots, 0, \Sigma^{(i)}, 0, \ldots, 0, \frac{1}{\varepsilon K}\Sigma^{(i)}]$ |
| $\boldsymbol{b}$ | A parameter used in linear programming formulation [Eq. 10]. Equals to $\mathbf{e}_j$ |
| $Z$ | A random variable follows the nonparanormal distribution |
| $\mathbf{S}$ | Correlation matrix of $Z$ |
| $\Sigma_{tot}$ | Defined in Sect. 5.1 |
| $\Omega_I$ | Defined in Sect. 5.1 |
| $\Omega_S^{tot}$ | Defined in Sect. 5.1 |
| $\Omega_{tot}$ | Defined in Sect. 5.1 |
| $X_{tot}$ | Defined in Sect. 5.1 |
| $I_K$ | Defined in Sect. 5.1 |
| $\sigma_{ij}$ | An entry of $\Sigma_{tot}$ |
| $\omega_{ij}$ | An entry of $\Omega_{tot}$ |
| $\omega_j$ | A column of $\Omega_{tot}$ |
| $\hat{\Omega}_{tot}$ | Estimated $\Omega_{tot}$ |
| $\Omega_{tot}^0$ | True $\Omega_{tot}$ |
| $\omega_j^0$ | A column of $\Omega_{tot}^0$ |
| $\hat{\Sigma}_{tot}$ | Estimated $\Sigma_{tot}$ |
| $\Sigma_{tot}^0$ | True $\Sigma_{tot}$ |
| $\hat{\omega}_j$ | A column of $\hat{\Omega}_{tot}$ |
| $\hat{\Omega}_{tot}^1$ | Solution of Eq. 15 |
| $\lambda_n$ | Hyper-parameter for the sparsity in Eq. 7 |
| $\varepsilon$ | Hyper-parameter balancing shared and individual part in Eq. 6 |
| $K$ | Total number of tasks |
| $p$ | Total number of features |
| $C$ | A constant in condition (C1) [Eq. 17] and (C2) [Eq. 18] |
| $q$ | A constant between 0 and 1 [Eq. 19] |
| $\eta$ | A constant between 0 and 0.25 in condition (C1) [Eq. 17] |
| $\gamma, \delta$ | Two constants in condition (C2) [Eq. (18)] |
| $M$ | A constant represents the upper bound in Eq. (19) |
| $s_0(p)$ | A constant represents the sparsity level of $\Omega$ in Eq. (19) |
| $\hat{\mathbf{B}}, \hat{\mathbf{B}}_I, \hat{\mathbf{B}}_S$ | Solution of Eq. (9) |
| $\hat{\boldsymbol{b}}_j^{(i)}$ | A column of $\hat{\mathbf{B}}_I$ |
| $\hat{\boldsymbol{b}}_j^S$ | A column of $\hat{\mathbf{B}}_I$ |
| $\tau_0$ | A constant in Theorem 4 |
| $C_4, C_5$ | Constants in Theorem 3 |
| $C_0, C_1$ | Constants in Theorem 4(a) |
| $\theta_0$ | Equals to $\max_{i,j,k} \hat{\Sigma}_{j,k}^{(i)}$ |
| $C_2, C_3$ | Constants in Theorem 4(b) |
| $\hat{\sigma}_{ij}$ | An entry of $\hat{\Sigma}_{tot}$ |
| $\sigma_{ij}^0$ | An entry of $\Sigma_{tot}^0$ |
| $C_{K1}, C_{K2}$ | Two constants used in the proof of Theorem 4(b) |
| $\mathbf{h}_j$ | $\hat{\omega}_j - \omega_j^0$ |
| $\mathbf{h}_j^1$ | $(\hat{\omega}_{ij} I\{|\hat{\omega}_{ij}| \geq 2t_n\}; 1 \leq i \leq p)^T - \omega_j^0$ |
| $\mathbf{h}_j^2$ | $\mathbf{h}_j - \mathbf{h}_j^1$ |
| $\bar{\mathbf{Y}}_{kij}$ | $\mathbf{X}_{ki}\mathbf{X}_{kj} I\{|\mathbf{X}_{ki}\mathbf{X}_{kj}| \leq \sqrt{n_{tot}/(\log Kp)^3}\} - \mathbb{E}\mathbf{X}_{ki}\mathbf{X}_{kj} I\{|\mathbf{X}_{ki}\mathbf{X}_{kj}| \leq \sqrt{n_{tot}/(\log Kp)^3}\}$ |
| $\check{\mathbf{Y}}_{kij}$ | $\mathbf{Y}_{kij} - \bar{\mathbf{Y}}_{kij}$ |
| $b_n$ | $\max_{i,j} \mathbb{E}|\mathbf{X}_{ki}\mathbf{X}_{kj}| I\{|\mathbf{X}_{ki}\mathbf{X}_{kj}| \leq \sqrt{n_{tot}/(\log Kp)^3}\}$ |
| $\mathbf{Y}_{kij}$ | $\mathbf{X}_{tot\,ki}\mathbf{X}_{tot\,kj} - \mathbb{E}\mathbf{X}_{tot\,ki}\mathbf{X}_{tot\,kj}$ |
| $\Phi^{(i)}$ | The inverse of $i$th correlation matrix for the nonparanormal case |

# 1 Introduction

Undirected graphical models (UGMs) provide a powerful tool for understanding statistical relationships among random variables. In a typical setting, we can represent the conditional dependency patterns among $p$ random variables $\{X_1, \ldots, X_p\}$ using an undirected graph $G = (V, E)$. $V$ includes $p$ nodes corresponding to the $p$ variables. $E$ denotes the set of edges describing conditional dependencies among the variables $\{X_1, \ldots, X_p\}$. If a pair of random variables is conditionally dependent given the rest of variables, there exists an edge in $E$ connecting the corresponding pair of nodes in $V$; otherwise, the edge is absent. Within the graphical model framework, the task of estimating such undirected graphs based on a set of observed data samples is called structure estimation or model selection. Much of the related literature has focused on estimating $G$ from a given data matrix $\mathbf{X}_{n \times p}$ (with $n$ observations across $p$ random variables) that are independently and identically drawn from $N_p(\mu, \Sigma)$. Here $N_p(\mu, \Sigma)$ represents a multivariate Gaussian distribution with mean vector $\mu$ ($\mu \in \mathbb{R}^p$) and covariance matrix $\Sigma$ (with size $p \times p$). Using the aforementioned $G$ to describe pairwise dependencies among $p$ variables for such a multivariate Gaussian distribution is called a Gaussian graphical model (GGM, e.g., Lauritzen 1996; Mardia et al. 1980). The inverse of the covariance matrix is called the precision matrix, $\Omega := \Sigma^{-1}$. Interestingly, GGM's conditional independence pattern corresponds to zeros of $\Omega$. This means that an edge does not connect $i$th node and $j$th node (i.e., conditionally independent) in GGM if and only if $\Omega_{ij} = 0$.

To achieve a consistent estimation of $G$, assumptions are usually imposed on the structure of $\Omega$. Most commonly, the graph sparsity assumption has been introduced by various estimators to derive sparse GGM (sGGM). The graph sparsity assumption corresponds to limiting the number of non-zero entries in the precision matrix $\Omega$, which leads to a combinational problem for structure estimation. Many classic GGM estimators use the $\ell_1$-norm to create convex relaxation of the combinatorial formulation. For instance, the popular estimator "graphical lasso" (GLasso) has considered maximizing $\ell_1$-penalized log-likelihood objective (Yuan and Lin 2007; Banerjee et al. 2008; Hastie et al. 2009; Rothman et al. 2008). More recently, Cai et al. (2011) proposed a constrained $\ell_1$-minimization formulation for estimating $\Omega$, known as the CLIME estimator. CLIME can be solved through column-wise linear programming and has shown favorable theoretical properties. Moreover, the nonparanormal graphical models (NGM) recently proposed by Liu et al. (2012) have extended sGGM to new distribution families. Both GGM and NGM belong to the general family of UGM (reviewed in Sect. 4).

This paper focuses on the problem of jointly estimating $K$ undirected graphical models from $K$ related multivariate sample blocks. Each sample block contains a different set of data observations on the same set of variables. This task is motivated by the fact that the past decade has seen a revolution in collecting large-scale heterogeneous data from many scientific fields like genetics and brain science. For instance, genomic technologies have delivered fast and accurate molecular profiling data across many cellular contexts (e.g. cell lines or cell stages) (ENCODE Project Consortium 2011). Many neuroimaging studies have collected functional measurements of brain regions across a cohort of multiple human subjects (Di Martino et al. 2014). Such networks can be concerned with identifying subject-specific variations across a population, where each individual is a unique context. For this type of data, understanding and quantifying context-specific variations across multiple graphs is a fundamental analysis task. Figure 1 provides a simple illustration (with two contexts) of the target problem. Interaction patterns that are activated only under a specific context can help to understand or to predict the importance of such a context (Ideker and Krogan 2012; Kelly et al. 2012).

Prior approaches for estimating UGMs from such heterogeneous data tend to either only estimate pairwise differential patterns between two graphs or jointly estimate multiple sGGMs toward a common graph pattern (reviewed in Sect. 4). The former strategy does not exploit the shared network structure across contexts and is not applicable for more than two contexts, leading to undesirable effects on the quality of the estimated networks. Conversely, the latter approach underestimates the network variability and makes implicit assumptions to minimize inter-context differences which are difficult to justify in practice. This is partly caused by the fact that relevant studies have mostly extended the "graphical lasso" (GLasso) estimator to multi-task settings and followed a penalized log-likelihood formulation [Eq. (2)]. Under the GLasso framework, however, explicitly quantifying the context-specific substructures involves a very challenging optimization task (explained in detail in Sect. 4).

This paper proposes a novel approach that uses a constrained $\ell 1$-minimization formulation for joint structure learning of multiple sparse GGMs or NGMs. We name the method SIMULE (detecting Shared and Individual parts of MULtiple graphs Explicitly), and include the following contributions:

- *Novel model* Using a constrained $\ell 1$ optimization strategy (Sect. 2), SIMULE extends CLIME to a multi-task setting. The learning step is solved efficiently through a formulation of multiple independent sub-problems of linear programming (Sect. 2.4) for which we also provide a parallel version of the learning algorithm. Compared with previous multi-task sGGM models, SIMULE can accurately quantify task-specific network varia-tions that are unique for each task. This also leads to a better generalization and benefits all the involved tasks.

– *Novel extension* Furthermore, since most real-world datasets do not follow the normality assumption, we extend SIMULE to Nonparanormal SIMULE (NSIMULE in Sect. 3) by learning multiple NGM under the proposed $\ell1$ constrained minimization. NSIMULE can deal with non-Gaussian data that follow the nonparanormal distribution (Liu et al. 2009), a much more generic data distribution family. Fitting NSIMULE is computationally as efficient as SIMULE.

– *Theoretical convergence rate* In Sect. 5, we theoretically prove that SIMULE and NSIMULE achieve a consistent estimation of the target (true) dependency graphs with a high probability at the rate $O(\log(Kp)/n_{tot})$. Here $n_{tot}$ represents the total number of samples from all tasks and $K$ describes the number of tasks. This proof also theoretically validates the benefit of learning multiple sGGMs jointly (Sect. 7), since the $O(\log(Kp)/n_{tot})$ convergence rate is better than learning multiple single-sGGMs separately at rate $O(\log p/n_i)$. $n_i$ represents the number of samples of $i$th task. Such an analysis hasn't been provided in any of the previous multi-sGGM studies.

– *Performance improvement* In Sect. 6 we show a strong improved performance of SIMULE and NSIMULE over multiple baseline methods on multiple synthetic datasets and two real-world multi-cell biomedical datasets. The proposed methods obtain better AUC and partial AUC scores on all simulated cases. On two real-world datasets, our methods find the most matches of variable interactions when using existing BioMed-databases for validation.

## 2 Method: SIMULE

Learning multiple UGM jointly is a task of interests in many applications. This paper tries to model and learn context-specific graph variations explicitly, because such variations can "fingerprint" important markers for fields like cognition (Ideker and Krogan 2012), physiology (Kelly et al. 2012) or pathology (Ideker and Krogan 2012; Kelly et al. 2012). We consider the general case of estimating $K$ graphical models from a $p$-dimensional aggregated dataset in the form of $K$ different data blocks.

In what follows, plain letters denote scalars. Uppercase and lowercase bold letters denote matrices and vectors respectively.[1] We denote $\mathbf{X}^{(i)}_{n_i \times p}$ as the $i$th data block (or data matrix). The total number of data samples uses notation $n_{tot} = \sum_{i=1}^{K} n_i$. The precision matrix uses notation $\Omega$ and the covariance matrix uses notation $\Sigma$. We denote the correlation matrix as $S$ and the inverse of correlation matrix as $\Phi$. The vector $\mathbf{e}_j = (0, \ldots, 1, \ldots, 0)^T$ denotes a basis vector in which only the $j$th entry is 1 and the rest are 0. For a $p$-dimensional data vector $\mathbf{x} = (x_1, x_2, \ldots, x_p)^T \in \mathbb{R}^p$, let $||\mathbf{x}||_1 = \sum_i |x_i|$ be the $\ell_1$-norm of $\mathbf{x}$, $||\mathbf{x}||_\infty = \max_i |x_i|$ be the $\ell_\infty$-norm of $\mathbf{x}$ and $||\mathbf{x}||_2 = \sqrt{\sum_i x_i^2}$ be the $\ell_2$-norm of $\mathbf{x}$. Similarly, for a matrix $\mathbf{X}$, let $||\mathbf{X}||_1 = \sum_{i,j} |\mathbf{X}_{i,j}|$ be the $\ell_1$-norm of $\mathbf{X}$ and $||\mathbf{X}||_\infty = \max_{i,j} |\mathbf{X}_{i,j}|$ be the $\ell_\infty$-norm of $\mathbf{X}$. $||\mathbf{X}||_2 = \sqrt{\lambda_{\max}(\mathbf{X})}$, here $\lambda_{\max}$ is the largest eigenvalue of $\mathbf{X}$. $||\mathbf{X}||_F = \sqrt{\sum_{i,j} \mathbf{X}_{i,j}^2}$ is the $F$-norm of $\mathbf{X}$. $||\mathbf{X}||_\mathbf{1} = \max_j \sum_i |\mathbf{X}_{ij}|$ is the matrix $\mathbf{1}$-norm of $\mathbf{X}$. $||\mathbf{X}^{(1)}, \mathbf{X}^{(2)}, \ldots, \mathbf{X}^{(K)}||_{1,p} = \sum_i ||\mathbf{X}^{(i)}||_p$ is the $\ell_{1,p}$-norm of $(\mathbf{X}^{(1)}, \mathbf{X}^{(2)}, \ldots, \mathbf{X}^{(K)})$. $\Omega \succ 0$ means that $\Omega$ is a positive definite matrix. $||(\Omega^{(1)}, \Omega^{(2)}, \ldots, \Omega^{(K)})||_{\mathcal{G},2} = $ $\sum_{j=1}^{p} \sum_{k=1}^{p} ||(\Omega_{j,k}^{(1)}, \Omega_{j,k}^{(2)}, \ldots, \Omega_{j,k}^{(i)}, \ldots, \Omega_{j,k}^{(K)})||_2 . ||(\Omega^{(1)}, \Omega^{(2)}, \ldots, \Omega^{(K)})||_{\mathcal{G},\infty} = \sum_{j=1}^{p} \sum_{k=1}^{p} ||(\Omega_{j,k}^{(1)}, \Omega_{j,k}^{(2)}, \ldots, \Omega_{j,k}^{(i)}, \ldots, \Omega_{j,k}^{(K)})||_\infty$.

---

[1] Following convention, $\Sigma, \Omega, \Phi, \beta, \mu, \theta$ and $I$ are not bold.

## 2.1 Background: CLIME for estimating sparse Gaussian graphical model

The CLIME estimator, short for constrained $\ell_1$ minimization method for inverse matrix estimation (Cai et al. 2011), uses an $\ell_1$ constrained optimization [as Eq. (1)] to estimate the precision matrix $\Omega$:

$$\underset{\Omega}{\operatorname{argmin}} ||\Omega||_1, \quad \text{subject to: } ||\Sigma\Omega - I||_\infty \leq \lambda \tag{1}$$

Here $\lambda > 0$ is the tuning parameter.

The idea comes from taking the first derivative of the objective function of "graphical Lasso" (GLasso) (Yuan and Lin 2007; Friedman et al. 2008). The original objective function (an $\ell_1$-penalized log-likelihood) is as follows,

$$\widehat{\Omega}_{glasso} = \underset{\Omega}{\operatorname{argmin}}\{-\log\det(\Omega)+ <\Omega, \Sigma> +\lambda||\Omega||_1\} \tag{2}$$

By taking the first derivative of Eq. (2) and setting it equal to zero, the solution $\widehat{\Omega}_{glasso}$ also satisfies:

$$\widehat{\Omega}_{glasso}^{-1} - \hat{\Sigma} = \lambda\widehat{Z} \tag{3}$$

where $\widehat{Z}$ is an element of the subdifferential $\partial ||\widehat{\Omega}_{glasso}||_1$. This leads to the CLIME estimator in Eq. (1). One of the best properties of CLIME is that it can be solved column by column separately. Suppose $\beta$ is one of the column vectors in the precision matrix $\Omega$. Instead of estimating the entire $\Omega$ all at once as in Eq. (1), we can estimate each column $\beta$ of $\Omega$ as follows:

$$\operatorname{argmin}||\beta||_1 \quad \text{subject to } ||\Sigma\beta - e_j||_\infty \leq \lambda$$

After we obtain an estimated $\widehat{\Omega}^1$ from Eq. (1), to maintain the symmetric property of the estimator, CLIME then uses the following operation,

$$\widehat{\Omega}_{ij} = \widehat{\Omega}_{ji} = \widehat{\Omega}_{ij}^1 \operatorname{sign}\left(\max\left(\left|\widehat{\Omega}_{ij}^1\right| - \left|\widehat{\Omega}_{ji}^1\right|, 0\right)\right) + \widehat{\Omega}_{ji}^1 \operatorname{sign}\left(\max\left(\left|\widehat{\Omega}_{ji}^1\right| - \left|\widehat{\Omega}_{ij}^1\right|, 0\right)\right)$$

Cai et al. (2011) has proved that CLIME achieves a consistent estimation of the true graph with a high probability at the rate $O(\log p/n)$.

## 2.2 Background: multi-task learning with task-shared and task-specific parameters

Multi-task learning (MTL) was initially proposed by Caruana (1997) to find common feature weights across multiple relevant tasks. If different tasks are sufficiently related, MTL can lead to a better generalization across all tasks. The specific MTL formulation we explore is suggested by MT-SVM (Evgeniou and Pontil 2004), which models multiple support vector machines (SVMs) through task-joint and task-specific parameters. When given $K$ related tasks in which each sample $(\boldsymbol{x}_i, y_i)$ belongs to exactly one of the tasks $\{1, \ldots, K\}$, MT-SVM (Evgeniou and Pontil 2004) learns $K$ distinct parameters $\boldsymbol{w}_1, \ldots, \boldsymbol{w}_K$ where each $\boldsymbol{w}_k$ is specifically dedicated for the task $k$. In addition, MT-SVM utilizes a global parameter $\boldsymbol{w}_0$ to capture the commonality among all the tasks. Each sample $\boldsymbol{x}$ is classified by $\hat{y}_i = \operatorname{sign}(\boldsymbol{x}_i^T(\boldsymbol{w}_0 + \boldsymbol{w}_k))$. Clearly, the vectors $\boldsymbol{w}_k$ are "small" when the tasks are similar to each other. In other words, we assume that the tasks are related in such a way that the true models are all close to some (shared) model $\boldsymbol{w}_0$. MT-SVM estimates all $\boldsymbol{w}_1, \ldots, \boldsymbol{w}_K$ and $\boldsymbol{w}_0$ simultaneously.

### 2.3 SIMULE: infer <u>S</u>hared and <u>I</u>ndividual parts of <u>MUL</u>tiple sGGM <u>E</u>xplicitly

Treating sparse GGM estimation from each data block as a single task, our main goal is to learn multiple sGGMs over $K$ tasks jointly, which can lead to a better generalization across all of the involved tasks (theoretically proven in Sect. 5).

Mirroring the strategy used by previous joint-estimators of multiple GLasso (described in Sect. 4), we achieve "multi-tasking" by summing up the CLIME estimators from each of the K tasks.

$$\hat{\Omega}^{(i)} = \underset{\Omega^{(i)}}{\operatorname{argmin}} \sum_i ||\Omega^{(i)}||_1$$

$$\text{Subject to: } ||\Sigma^{(i)}\Omega^{(i)} - I||_\infty \le \lambda_{n_{tot}}, \quad i = 1, \ldots, K \qquad (4)$$

Here $n_{tot} = \sum_{i=1}^{K} n_i$ represents the total number of data samples used in all tasks. It is worth mentioning that we select the hyperparameter by considering the sample size, i.e., using $\lambda_{n_{tot}}$ instead of $\lambda$ (based on two recent studies (Negahban et al. 2009; Yang et al. 2014), details in Sect. 6). To simplify notations, we use $\lambda_n$ instead of $\lambda_{n_{tot}}$ in the rest of this section.

Then following the MTL formulation of MT-SVM, we simply model each sGGM network as

$$\Omega^{(i)} = \Omega_I^{(i)} + \Omega_S, \qquad (5)$$

where $\Omega_S$ is the shared pattern among all graphs and $\Omega_I^{(i)}$ represents the individual part specific for $i$th graph (see Fig. 1 for a simple illustration). Furthermore, we assume both the shared and the individual parts of each $\Omega^{(i)}$ should be sparse and convert this assumption into the following formulation,

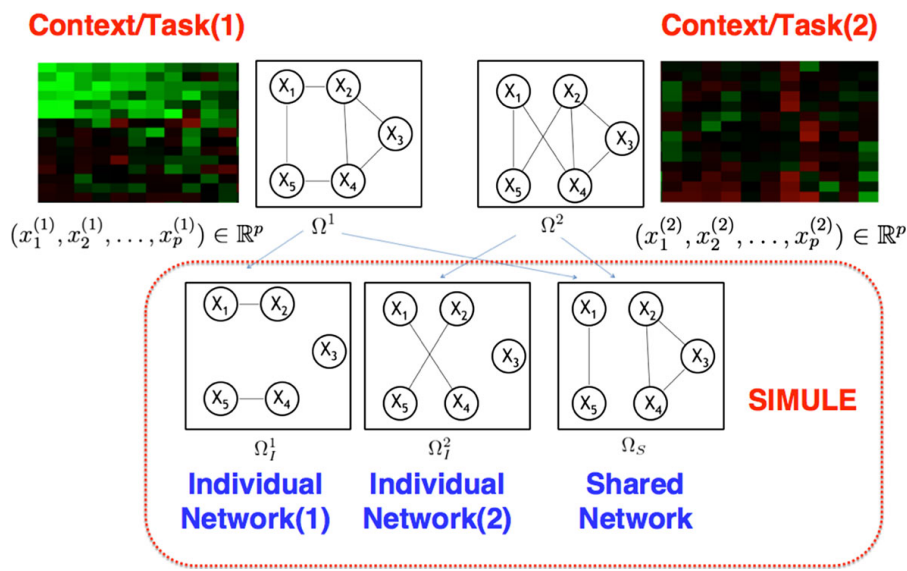

**Fig. 1** A simple example of the target problem. We are trying to learn (1) individual parts that are specific to each task itself (*two left lower subfigures*) and (2) at the same time, the shared part that is common for all tasks (*the right lower subfigure*)

$$\left|\left|\Omega_I^{(i)}\right|\right|_1 + \varepsilon ||\Omega_S||_1 \qquad (6)$$

Here we assume $\varepsilon > 0$. This hyper-parameter is chosen according to applications. It reflects the difference of sparsity level between the shared part and the context-specific parts. For example, due to evolutionary stability and/or system stability, we usually assume that the shared part of gene networks is more dense than individual subgraphs of each cell context. Accordingly, Sect. 6 uses $\varepsilon < 1$ for two real-world molecular expression datasets.

Now we replace the objective in Eq. (4) with the new objective from Eq. (6) and plug Eq. (5) into the constraints of Eq. (4). This gives the following novel formulation of SIMULE:

$$\hat{\Omega}_I^{(1)}, \hat{\Omega}_I^{(2)}, \ldots, \hat{\Omega}_I^{(K)}, \hat{\Omega}_S = \underset{\Omega_I^{(i)}, \Omega_S}{\mathrm{argmin}} \sum_i \left|\left|\Omega_I^{(i)}\right|\right|_1 + \varepsilon K ||\Omega_S||_1$$

$$\text{Subject to: } ||\Sigma^{(i)}(\Omega_I^{(i)} + \Omega_S) - I||_\infty \le \lambda_n, \quad i = 1, \ldots, K \qquad (7)$$

In Sect. 5, we theoretically prove that those $\{\Omega^{(i)} | i = 1, \ldots, k\}$ estimated from Eq. (7) are positive definite and converge to the true precision matrices with a high probability.

Equation (7) can be solved column by column without influencing the resulting solution. For each column, we solve the following optimization:

$$\underset{\beta^{(i)}, \beta^s}{\mathrm{argmin}} \sum_i ||\beta^{(i)}||_1 + \varepsilon K ||\beta^s||_1$$

$$\text{Subject to: } ||\Sigma^{(i)}(\beta^{(i)} + \beta^s) - e_j||_\infty \le \lambda_n, \quad i = 1, \ldots, K \qquad (8)$$

$\beta^{(i)}$ is one of the column vectors in the individual part $\Omega_I^{(i)}$ of $i$th graph (we take out the subscript $I$ in $\beta^{(i)}$ to simplify notations). $\beta^s$ is the corresponding column in the shared part $\Omega_S$.

Equivalently, SIMULE solves the following optimization, for each column:

$$\underset{\beta^{(i)}, \beta^s}{\mathrm{argmin}} \left|\left|\left[\beta_{p \times 1}^{(1)}, \ldots, \beta_{p \times 1}^{(i)}, \ldots, \beta_{p \times 1}^{(K)}, \varepsilon K \beta_{p \times 1}^s\right]_{p \times (K+1)}\right|\right|_1,$$

Subject to:

$$||[0, \ldots, 0, \Sigma^{(i)}, 0, \ldots, 0, \Sigma^{(i)}]_{p \times (K+1)p}$$
$$[\beta^{(1)^T}, \ldots, \beta^{(i)^T}, \ldots, \beta^{(K)^T}, (\beta^s)^T]_{(K+1)p \times 1}^T$$
$$- e_j|_\infty \le \lambda_n, \quad i = 1, \ldots, K. \qquad (9)$$

## 2.4 Optimization

By simplifying notations, Eq. (9) can be rewritten as

$$\underset{\theta}{\mathrm{argmin}} ||\theta||_1$$

$$\text{Subject to: } |\mathbf{A}^{(i)}\theta - b|_\infty \le c, \ i = 1, \ldots, K$$

$$\text{Where } \mathbf{A}^{(i)} = \left[0, \ldots, 0, \Sigma^{(i)}, 0, \ldots, 0, \frac{1}{\varepsilon K}\Sigma^{(i)}\right],$$

$$\theta = \left[\beta^{(1)^T}, \ldots, \beta^{(i)^T}, \ldots, \beta^{(K)^T}, \varepsilon K (\beta^s)^T\right]^T, \ \boldsymbol{b} = \mathbf{e}_j, c = \lambda_n \qquad (10)$$

**Algorithm 1** Jointly estimate $\underline{S}$hared and $\underline{I}$ndividual Parts of $\underline{Mul}$tiple sGGM $\underline{E}$xplicitly(SIMULE)

---

**Input:** Data sample matrix $\mathbf{X}^{(i)}$ ( $i = 1$ to $K$ ), regularization hyperparameter $\lambda_n$, hyperparameter $\varepsilon$ and $\mathbf{LP}(.)$ (a linear programming solver)

**Output:** Shared part $\Omega_S$ and context-specific parts $\Omega_I^{(i)}$ (size $p * p$), ($i = 1$ to $K$)

1: **for** $i = 1$ to $K$ **do**
2:    Initialize $\Sigma^{(i)} = \frac{1}{n_i - 1} \sum_{s=1}^{n_i} (\mathbf{X}_{s,}^{(i)} - \hat{\mu}^{(i)})(\mathbf{X}_{s,}^{(i)} - \hat{\mu}^{(i)})^T$ (the sample covariance matrix of $\mathbf{X}^{(i)}$)
3:    Initialize $\Omega_I^{(i)} = \mathbf{0}_{p \times p}$
4:    Initialize $\mathbf{A}^{(i)} = [0, \ldots, 0, \Sigma^{(i)}, 0, \ldots, 0, \frac{1}{\varepsilon K} \Sigma^{(i)}]$
5: **end for**
6: Initialize $\Omega_S = \mathbf{0}_{p \times p}$
7: $c = \lambda_n$
8: **for** $j = 1$ to $p$ **do**
9:    $\boldsymbol{b} = \mathbf{e}_j$
10:    $\theta = \mathbf{LP}(\mathbf{A}^{(i)}, \boldsymbol{b}, c)$ where $i = 1, \ldots, K$ and $\mathbf{LP}(.)$ solves Eq. (11)
11:    **for** $i = 1$ to $K$ **do**
12:       $\Omega_{I\ ,j}^{(i)} = \theta_{((i-1)p+1):ip}$
13:    **end for**
14:    $\Omega_{S,j} = \theta_{(Kp+1):(K+1)p}$
15: **end for**

---

Through relaxation, we convert Eq. (10) to the following linear programming formulation in Eq. (11),

$$\operatorname*{argmin}_{\mathbf{u}_j, \theta} \sum_{j=1}^{(K+1)p} \mathbf{u}_j$$

subject to:

$$-\theta_j \leq \mathbf{u}_j, \quad j = 1, \ldots, (K+1)p$$
$$\theta_j \leq \mathbf{u}_j, \quad j = 1, \ldots, (K+1)p$$
$$-\mathbf{A}_{k,}^{(i)T} \theta + \boldsymbol{b}_k \leq c, \quad k = 1, \ldots, p, i = 1, \ldots, K$$
$$\mathbf{A}_{k,}^{(i)T} \theta - \boldsymbol{b}_k \leq c, \quad k = 1, \ldots, p, i = 1, \ldots, K \tag{11}$$

Here a set of $\mathbf{u}_j$ are the slack variables, $\mathbf{A}_{k,}^{(i)}$ means the $k$th row of $\mathbf{A}^{(i)}$ and $\boldsymbol{b}_k$ is the $k$th entry of $b$. The pseudo code of SIMULE is summarized in Algorithm 1. Following CLIME, then we apply the same symmetric operators on $\{\Omega^{(i)} = \Omega_S + \Omega_I^{(i)}\}$ obtained from Algorithm 1. Section 5 proves that $\Omega^{(i)}$ converges to the true one with a optimal convergence rate.

To solve Eq. (11), we follow the primal dual interior method (Boyd and Vandenberghe 2004) that has also been used in the Dantzig selector for the task of regression (Candes and Tao 2007). Other strategies can be used to solve this linear programming, such as the one used in Pang et al. (2014).

## 2.5 Parallel version of SIMULE

Algorithm 1 can easily be revised into a parallel version. Essentially we just need to revise the "For loop" of step 8 in Algorithm 1 into, for instance, "column per machine" or "column per core". Since the calculation of each column is independent from the other columns, this parallel variation will obtain the same solution as Algorithm 1 at a better speed. Section 6 shows the speed improvements of SIMULE in a multi-core setting versus a single-core setting.

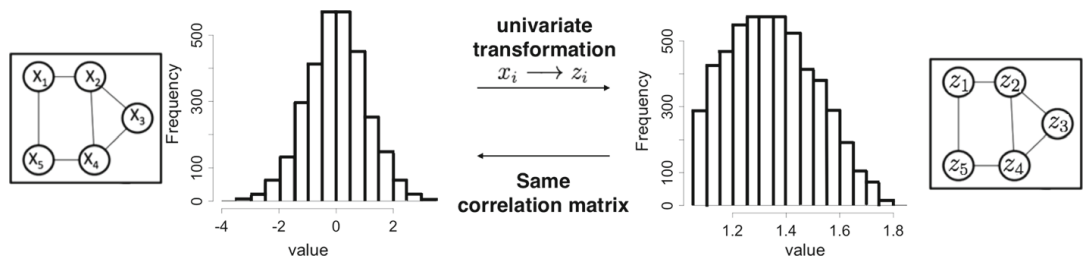

**Fig. 2** A simple example showing nonparanormal graphical model and its unobserved latent Gaussian graphical model. The *leftmost sub-figure* shows $X \sim N_p(\mu, \mathbf{S})$. The *rightmost sub-figure* shows $Z \sim NPN_p(\mu, \mathbf{S}; f_1, \ldots, f_p)$. The *right distribution graph* shows the histogram of one feature $\mathbf{z}_i$ (one TF variable, data details in Sect. 6.6) from a real TF binding dataset. The *left histogram graph* shows the distribution of a log-transformation of the same feature ($\mathbf{z}_i$). Because we can clearly see that the *left histogram* roughly follows a Gaussian distribution, this indicates $\mathbf{z}_i$ follows a nonparanormal distribution. This shows the need to extend SIMULE to the nonparanormal distribution that is a strict superset of the Gaussian distribution

## 3 Method variation: nonparanormal SIMULE (NSIMULE)

Though sGGM is powerful, its normality assumption is commonly violated in real applications. For instance, for the TF ChIP-Seq data analyzed in Sect. 6.6, the histogram of one of its TF variables is clearly not following Gaussian distribution (across samples, shown as the right distribution graph in Fig. 2). After a univariate log-transformation of the same feature, we obtain its distribution histogram as the left graph in Fig. 2. The transformed data samples are approximately normally distributed. This motivates us to adopt a more generalized UGM (recently proposed in Liu et al. 2009) to overcome the limitation of sGGM. This so-called "nonparanormal graphical model" (Liu et al. 2009) assumes that data samples follow a multivariate nonparanormal distribution, which is a strict superset of the Gaussian distribution. We extend SIMULE to the nonparanormal family and name this novel variation NSIMULE. NSIMULE learns to fit multiple NGMs jointly through modeling task-shared and task-specific parameters explicitly.

### 3.1 Background: nonparanormal graphical model

A random variable $Z = (Z_1, \ldots, Z_p)^T$ is said to follow a nonparanormal distribution

$$Z \sim NPN_p(\mu, \mathbf{S}; f_1, \ldots, f_p)$$

if and only if there exists a set of univariate strictly increasing transformations $f = \{f_j\}_{j=1}^{p}$ such that:

$$f(Z) = (f_1(Z_1), \ldots, f_p(Z_p))^T := X \sim N(\mu, \mathbf{S})$$

Figure 2 shows a simple example of NGM (inside the rightmost sub-figure) and its unobserved latent Gaussian graphical model (inside the leftmost sub-figure). Assume that we are given a dataset including $n$ observations that are independently and identically drawn from $NPN_p(\mu, \mathbf{S}; f_1, \ldots, f_p)$, a multivariate nonparanormal distribution. The conditional independence graph $G$ among $Z_i$ variables can be modeled with a corresponding NGM (Liu et al.

2009). The graph structure of NGM is encoded through the sparsity pattern of the inverse covariance matrix $\Phi := \mathbf{S}^{-1}$, where $\mathbf{S}$ denotes the covariance matrix of $NPN_p$.

### 3.2 Background: estimate S through rank-based measures of correlation matrix $\mathbf{S_0}$

Since the direct estimation of covariance matrix $\mathbf{S}$ is difficult in nonparanormal distribution, recent studies have proposed an efficient nonparametric estimator (Liu et al. 2009) for $\mathbf{S}$. This estimator is derived from the correlation matrix $\mathbf{S}_0$. Because the covariance matrix $\mathbf{S} = diag(\mathbf{S}_i)\mathbf{S}_0 diag(\mathbf{S}_i)$, $\mathbf{S}^{-1} = diag(\mathbf{S}_i)^{-1}\mathbf{S}_0^{-1}diag(\mathbf{S}_i)^{-1}$. Here $\mathbf{S}_i = \sqrt{Cov(Z_i, Z_i)}$ and $diag(\mathbf{S}_i) = diag(\mathbf{S}_1, \mathbf{S}_2, \ldots, \mathbf{S}_p)$. Therefore, the inverse of correlation matrix ($\mathbf{S}_0^{-1}$) and the inverse of covariance matrix ($\mathbf{S}^{-1}$) have the same nonzero and zero entries. Based on this observation, Liu et al. (2009) proposed a nonparametric method to estimate the correlation matrix $\mathbf{S}_0$, instead of estimating the covariance matrix $\mathbf{S}$ for the purpose of structure inference.

In Liu et al. (2009) the authors proposed using the population Kendall's tau correlation coefficients $\tau_{jk}$ to estimate $\mathbf{S}_0$, based upon the explicit relationship between this rank-based measure $\tau_{jk}$ and the correlation measure $(\mathbf{S}_{jk})_0$ for a given nonparanormal dataset $Z \sim NPN_p(\mu, \mathbf{S}, f_1, \ldots, f_p)$ (discussed in Liu et al. 2012). Figure 2 presents the simple relationship between $Z \sim NPN_p(\mu, S; f_1, \ldots, f_p)$ and its latent $X \sim N(\mu, S)$. To simplify notations, we use $\mathbf{S}$ to represent the correlation matrix for the remainder of this paper.

**Theorem 1** *Given $Z \sim NPN_p(\mu, S, f_1, \ldots, f_p)$, a nonparanormal distribution, we have that*

$$\mathbf{S}_{jk} = sin\left(\frac{\pi}{2}\tau(Z_j, Z_k)\right). \tag{12}$$

where the Kendall's tau can be estimated as:

$$\hat{\tau}_{jk} = \frac{1}{n(n-1)} \sum_{1 \leq i \leq i' \leq n} \text{sign}\left(\left(\mathbf{z}_j^i - \mathbf{z}_j^{i'}\right)\left(\mathbf{z}_k^i - \mathbf{z}_k^{i'}\right)\right)$$

*Proof* The proof is provided in Liu et al. (2009). □

Therefore, the correlation matrix $\mathbf{S}$ can be estimated as:

$$\widehat{\mathbf{S}}_{jk} = sin\left(\frac{\pi}{2}\hat{\tau}_{jk}\right).$$

We can then plug in the estimated $\widehat{\mathbf{S}}$ for learning the dependency graph structure in the corresponding NGM.

### 3.3 NSIMULE: nonparanormal SIMULE

We can now substitute each sample covariance matrix $\hat{\Sigma}^{(i)}$ used in Eq. (9) from each task with its corresponding correlation matrix $\mathbf{S}^{(i)}$ as estimated above. The rest of the computations are the same as SIMULE. We refer to this whole process as NSIMULE. It estimates multiple different, but related, sparse Nonparametric Graphical Models (sNGM) through shared and task-specific parameter representations.

**Theorem 2** *If $X, Y$ are two independent random variables and $f, g : \mathbb{R} \to \mathbb{R}$ are two measurable functions, then $f(X)$ and $g(Y)$ are also independent.*

Through the above theorem, the monotone functions $f$ in $NPN_p$ will not change the conditional dependency among variables. As proved in Liu et al. (2009), the conditional dependency network among the latent Gaussian variables $X$ (in this $NPN_p$) is the same as the conditional dependency network among the nonparanormal variables $Z_i$, with a parametric asymptotic convergence rate. Therefore, we can use the estimated correlation matrices $\mathbf{S}^{(i)}$ for the joint network inference of multiple sNGMs in SIMULE. This is also because we have shown that the inverse of the correlation matrix and the inverse of the covariance matrix share the same nonzero and zero patterns.

## 4 Related work

### 4.1 Connecting to past multi-sGGM studies

Sparse GGM is an extremely active topic in the recent literature including notable studies like Wainwright and Jordan (2006) and Banerjee et al. (2008). We can categories single-task sGGM estimators into three groups: (a) penalized likelihood (GLasso), (b) neighborhood approach and (c) CLIME estimator.

Most previous methods that estimate multiple sGGMs jointly (on the same set of variables from aggregated data samples) can be formulated as:

$$\operatorname*{argmin}_{\Omega^{(i)}>0} \sum_i (-L(\Omega^{(i)}) + \lambda_1 \sum_i ||\Omega^{(i)}||_1$$
$$+ \lambda_2 P(\Omega^{(1)}, \Omega^{(2)}, \dots, \Omega^{(K)}) \tag{13}$$

where $\Omega^{(i)}$ denotes the precision matrix for the $i$th task. $L(\cdot)$ represents log-likelihood or pseudo-likelihood function. $\sum_i ||\Omega^{(i)}||_1$ adds sparsity constraints on each task. $P(\Omega^{(1)}, \Omega^{(2)}, \dots, \Omega^{(K)})$ enforces certain joint properties among the tasks (like using group sparsity to enforce similarity among tasks).

We choose the most relevant three studies as our baselines in the experiments: (a) Fused Joint graphical lasso (JGL-fused) (Danaher et al. 2013), (b) Group Joint graphical lasso (JGL-group) (Danaher et al. 2013) and (c) SIMONE (Chiquet et al. 2011). JGL-fused and JGL-group are based on the popular "graphical lasso" estimator (Friedman et al. 2008; Yuan and Lin 2007); [using $L(\Omega) = (\log det(\Omega) - < \Sigma, \Omega >)$ in Eq. (13)]. SIMONE (Chiquet et al. 2011) follows neighborhood-selection based estimator. It can be viewed as using a pseudo-likelihood approximation instead of the full likelihood as $L(\Omega)$ in Eq. (13).[2]

The general purpose of the penalty function $P(\Omega^{(1)}, \Omega^{(2)}, \dots, \Omega^{(K)})$ in Eq. (13) is to push the inference of multiple graphs toward a common pattern. Table 1 provides a representative list of penalty functions that have been used previously for the multi-sGGM setting. For example, JGL uses the fused norm (the 1st row of Table 1) to penalize the difference between two graphs with resulting variation named as JGL-fused. JGL-group uses a $\{\mathcal{G}, 2\}$ norm that pushes multiple graphs to have the same sparsity patterns (the 2nd row of Table 1). SIMONE provides a novel penalty proposed by the authors (shown as in the 3rd row of Table 1) to enforce similar sparsity pattern on multiple graphs.

In addition to these three baselines, a number of recent studies also perform multi-task learning of sGGM (Honorio and Samaras 2010; Guo et al. 2011; Zhang and Wang 2012; Zhang and Schneider 2010; Zhu et al. 2014). They all follow the same formulation as Eq. (13)

---

[2] JGL (Danaher et al. 2013) and SIMONE (Chiquet et al. 2011) are the two most-cited joint GGM estimators in the literature.

**Table 1** A list of representative multi-sGGM methods and the second penalty functions they have used

| | References | Penalty function $P(\Omega^{(1)}, \Omega^{(2)}, \ldots, \Omega^{(K)}) =$ |
|---|---|---|
| (1) | JGL-Fused (Danaher et al. 2013) | $\sum_{ij, i>j} \|\Omega^{(i)} - \Omega^{(j)}\|_1$ |
| (2) | JGL-Group (Danaher et al. 2013) | $\|\Omega^{(1)}, \Omega^{(2)}, \ldots, \Omega^{(K)}\|_{\mathcal{G}, 2}$ |
| (3) | SIMONE (Chiquet et al. 2011) | $\sum_{i \neq j} \left( \left( \sum_{k=1}^{T} \left( \Omega_{ij}^{(k)} \right)_{+}^2 \right)^{\frac{1}{2}} \right) + \left( \left( \sum_{k=1}^{K} \left( -\Omega_{ij}^{(k)} \right)_{+}^2 \right)^{\frac{1}{2}} \right)$ |
| (4) | Node JGL (Mohan et al. 2013) | $\sum_{ij, i>j} RCON(\Omega^{(i)} - \Omega^{(j)})$ |
| (5) | JEM-GM (Guo et al. 2011) | $\sum_{k=1}^{K} w_k \|\Omega^{(k)}\|_1$ |
| (6) | MTL-GGM (Honorio and Samaras 2010) | $\|\Omega^{(1)}, \Omega^{(2)}, \ldots, \Omega^{(K)}\|_{\mathcal{G}, \infty}$ |
| (7) | CSSL-GGM (Hara and Washio 2013) | $\|\Omega_S\|_1 + \|\Omega_I^{(1)}, \Omega_I^{(2)}, \ldots, \Omega_I^{(K)}\|_{1, p}$ |

but explore a different second penalty-function $P(\Omega^{(1)}, \Omega^{(2)}, \ldots, \Omega^{(K)})$. As an example, Node-based JGL proposed a novel penalty, namely $RCON$ (Mohan et al. 2013) (shown as the 4th row of the Table 1) or "row-column overlap norm" for capturing special relationship among graphs. In two recent works, the penalty function at the 5th row is Table 1 has been used by Guo et al. (2011) and the penalty function at the 6th row of Table 1 has been used by Honorio and Samaras (2010).

Furthermore, there exist studies that explore similar motivations as ours when learning multiple GGM models from data. (a) Han et al. (2013) proposed to estimate a population graph from multi-block data using a so-called "median-graph" idea. It is conceptually similar to $\Omega_S$. However, they do not have $\Omega_I^{(i)}$ to model individual parts that are specific to each task. (b) Another recent study, CSSL-GGM Hara and Washio (2013) also tried to model both the shared and individual substructures in multi-sGGMs. Different from ours, their formulation is within the penalized likelihood framework as Eq. (13). They used $\ell 1,p$ norm (see last row of Table 1) to regularize the task-specific parts, while SIMULE uses $\ell 1$ norm instead in Eq. (6). The $\ell 1,p$ norm pushes the individual parts of multiple graphs to be similar which is contradictory to the original purpose of these parameters.[3] (c) More recently, Monti et al. (2015) proposed to learn population and subject-specific brain connectivity networks via a so-called "Mixed Neighborhood Selection" (MSN) method. Following the neighborhood selection framework (Meinshausen and Bühlmann 2006), for each node $v$, MSN tried to learn the neighborhood of each $v$. Similar to SIMULE, they estimated the neighborhood edges of a given node $v$ in the $i$-task as $\beta^v + \widetilde{b}^{(i),v}$. Here $\beta^v$ represents the neighbor in the shared part and $\widetilde{b}^{(i),v}$ represents the neighbors that are specific to the $i$th graph. Since MSN is specially designed for brain imaging data, it assumes each individual graph is generated by random effects, i.e., $\widetilde{b}^{(i),v} \sim N(0, \Phi^v)$. SIMULE does not have such strong assumptions on either task-specific or task-shared substructures. Our model is more general while MSN is designed for brain imaging data. (d) Another line of related studies (Liu et al. 2013; Sugiyama et al. 2013; Fazayeli and Banerjee 2016) prosed density-ratio based strategies to estimate a differential graph between two graphs. Even though this group of methods can handle the unbalance dataset (i.e., the numbers of samples in two datasets are quite different), they can only capture the difference between two graphs ($K = 2$). SIMULE does not have such a limitation on the number of tasks. (e) Moreover, several loosely related studies exist in settings different from ours. For example, for handling high-dimensional time series data a few recent papers have considered exploring multiple sGGMs by modeling relationships among networks; e.g., Kolar et al. (2010), Qiu et al. (2013).

---

[3] We can not find CSSL-GGM implementation, therefore can not include it as a baseline.

### 4.2 Biasing covariance matrices with SIMONE-I and SIMULE-I

The SIMONE package (Chiquet et al. 2011) has introduced "intertwined Lasso" (named as "SIMONE-I" in the rest of paper) that takes the perspective of sharing the information among covariance matrices. More specifically, this variation averages each task's sample covariance matrix with a global empirical covariance matrix obtained from the whole dataset. Motivated by SIMONE-I, we extend SIMULE with a similar strategy that revises each task's sample covariance matrix with $\tilde{\Sigma}^{(i)} = \alpha \Sigma^{(i)} + (1 - \alpha) n_{tot}^{-1} \sum_{t=1}^{K} n_t \Sigma^{(t)}$. Here $\alpha = 0.5$. This variation is referred as "SIMULE-I" for the rest of this paper. We report experimental results from both SIMONE-I and SIMULE-I in Sect. 6.

Fan et al. (2014) have pointed out that sample covariance matrix is inconsistent under a high-dimensional setting. There exists a huge body of previous literature for covariance matrices estimation. Roughly relevant studies can be grouped into three types: (a) Sparse covariance matrix estimation, including studies like hard-thresholding (Lam and Fan 2009), soft-thresholding (Tibshirani 1996), smoothly clipped absolution deviation (SCAD, Fan et al. 2013), and minimax concavity penalties (Zhang 2010). These methods, though simple, do not guarantee the positive definiteness of the estimated covariance matrix. (b) Positive definite sparse covariance matrix estimation that introduces penalty functions on the eigenvector space. Popular studies include Antoniadis and Fan (2011), Liu et al. (2014), Levina et al. (2008), Rothman (2012). (c) Factor-model based covariance matrix estimation like POET (Fan et al. 2013) that applies soft-thresholding to the residual space obtained after applying an approximate factor structure on the estimated covariance matrix. Most of these studies can be used to extend SIMULE. We leave a more thorough study of such combinations as future work.

### 4.3 Penalized log-likelihood for SIMULE

Comparing Eq. (9) of SIMULE with Eq. (13), previous multi-sGGM approaches mostly relied on penalized log-likelihood functions for learning. We have also considered extending penalized log-likelihood method such as the "graphical lasso" estimator into our MTL setting: $\Omega^{(i)} = \Omega_I^{(i)} + \Omega_S$. However, it is hard to deal with the log determinant term $logdet(\Omega_I^{(i)} + \Omega_S)$ in optimization. This is because $\frac{\partial logdet(\mathbf{X+Y})}{\partial \mathbf{X}} = (\mathbf{X+Y})^{-1}$ and it is difficult to calculate the inverse of $\mathbf{X} + \mathbf{Y}$. There is no closed form for the inverse of the sum of two matrices, except for certain special cases. From the perspective of optimization, it is hard to directly use methods like coordinate descent for learning such a model due to this first derivative issue. The authors of CSSL-GGM (Hara and Washio 2013) handled the issue by using Alternating Direction Method of Multipliers (ADMM) . This optimization is much more complicated than SIMULE. Another package MNS (Monti et al. 2015) tackled this issue partly through adding latent variables. It assumes that each individual part $\tilde{\boldsymbol{b}}^{(i),v} \sim N(0, \Phi^v)$, in which $\Phi^v$ is a latent variable learned by the EM algorithm.

**Table 2** Two categories of relevant studies: learning with "penalized log-likelihood" or learning with "$\ell 1$ constrained-optimization"

| Tasks | Penalized likelihood | $\ell 1$ constrained-optimization |
|---|---|---|
| High dimensional linear regression | Lasso: $\text{argmin}_\beta \lVert \mathbf{Y} - \beta \mathbf{X} \rVert_F + \lambda \lVert \beta \rVert_1$ | Dantzig selector: $\text{argmin}_\beta \lVert \beta \rVert_1$ subject to : $\lVert \mathbf{X}^T \mathbf{y} - \mathbf{X}^T \mathbf{X} \beta \rVert_\infty \leq \lambda$ |
| sGGM | GLasso: $\text{argmin}_{\Omega \geq 0} - log det(\Omega) + <\Omega, \Sigma> + \lambda \lVert \Omega \rVert_1$ | CLIME: $\text{argmin}_{\Omega \geq 0} \lVert \Omega \rVert_1$ subject to: $\lVert \Omega \Sigma - I \rVert_\infty \leq \lambda$ |
| Multi-task learning of sGGM | Different penalty: $\text{argmin}_{\Omega^{(i)} > 0} \sum_i (-L(\Omega^{(i)}) + \lambda_1 \sum_i \lVert \Omega^{(i)} \rVert_1 + \lambda_2 P(\Omega^{(1)}, \Omega^{(2)}, \dots, \Omega^{(K)}))$ | Our SIMULE: $\text{argmin}_{\Omega_I^{(i)}, \Omega_S} \sum_i \left\lVert \Omega_I^{(i)} \right\rVert_1 + \varepsilon K \lVert \Omega_S \rVert$ subject to: $\lVert \Sigma^{(i)} (\Omega_I^{(i)} + \Omega_S) - I \rVert_\infty \leq \lambda_n, \ i = 1, \dots, K$ |

## 4.4 Relevant studies using $\ell_1$ optimization with constraints

The formulation of using $\ell_1$ based objective with constraints has been explored by other tasks before. (a) For example, the Dantzig estimator (Candes and Tao 2007) uses the following $\ell_1$ optimization with constraints for high-dimensional linear regression:

$$\underset{\beta}{\text{argmin}} \lVert \beta \rVert_1 \quad \text{subject to: } \lVert \mathbf{X}^T \mathbf{y} - \mathbf{X}^T \mathbf{X} \beta \rVert_\infty \leq \lambda.$$

This is in contrast to the lasso estimator using the penalized log-likelihood formulation for the high-dimensional regression:

$$\underset{\beta}{\text{argmin}} \lVert \mathbf{Y} - \beta \mathbf{X} \rVert_F + \lambda \lVert \beta \rVert_1$$

(b) For the case of single-task sGGM, the CLIME estimator (Cai et al. 2011) infers precision matrix through solving Eq. (1): $\ell 1$ optimization with constraints (and further derived into a linear programming formulation). This has been shown to provide more favorable theoretical properties than penalized likelihood approaches like GLasso [solving Eq. (2)]. (c) Based on our knowledge, the proposed SIMULE model is the first study to use constrained $\ell 1$ optimization for the case of multi-tasking sGGM. Table 2 summarizes relevant studies from the perspective of two different optimization formulations.

## 4.5 Optimization and computational concerns

Furthermore, a number of papers have focused on proposing ways to improve the performance of computation and data storage when estimating sGGMs. For example, the BigQUIC algorithm (Hsieh et al. 2011) aims at an asymptotic quadratic optimization when estimating sGGM. The authors of the Node-based JGL method (Mohan et al. 2013) proposed a block-separate method to improve the computational performance of multi-sGGMs. Our method can be extended to a parallel setting, since it can be naturally decomposed into per column based optimization. Node-based JGL (Mohan et al. 2013) has also used the ADMM optimization algorithm for learning. Compared to linear-programming-based SIMULE, Mohan et al. (2013) has two main disadvantages: (a) a large number of complex optimizations are required to be solved, and (b) the condition of convergence is unspecified. SIMULE makes use of an efficient algorithm, is proven to achieve good numerical performance, and ensures convergence (see Sect. 5).

Furthermore, since imposing an $\ell_1$ penalty on the model parameters formulates the structure learning of UGMs as a convex optimization problem, this strategy has shown successful results for modeling continuous data with GGM or NGM and discrete data with the pairwise Markov Random Fields (MRFs) (Friedman et al. 2008; Höfling and Tibshirani 2009). However, the discrete case of pairwise MRF is much harder because of the potentially intractable normalizing constant and also the possibility that each edge may have multiple parameters. One more complicating factor about structure learning is that the pairwise assumption might need a few exceptions, like searching for higher-order combinatorial interactions in recent studies like Schmidt and Murphy (2010), Buchman et al. (2012). (Detailed descriptions of related work are not included due to space limitation.)

## 5 Theoretical Analysis

### 5.1 Theoretical Analysis for Basic SIMULE

This section proves the theoretical properties of the SIMULE estimator. To present our analysis more concisely, we reformulate our model into a more general presentation as follows:

$$\Sigma_{tot} := \begin{pmatrix} \Sigma^{(1)} & 0 & \cdots & 0 \\ 0 & \Sigma^{(2)} & \cdots & 0 \\ \vdots & \vdots & \ddots & \vdots \\ 0 & 0 & \cdots & \Sigma^{(K)} \end{pmatrix} = (\sigma_{ij})_{Kp \times Kp}$$

$$\Omega_I := \begin{pmatrix} \Omega_I^{(1)} & 0 & \cdots & 0 \\ 0 & \Omega_I^{(2)} & \cdots & 0 \\ \vdots & \vdots & \ddots & \vdots \\ 0 & 0 & \cdots & \Omega_I^{(K)} \end{pmatrix}$$

$$\Omega_S^{tot} := \begin{pmatrix} \Omega_S & 0 & \cdots & 0 \\ 0 & \Omega_S & \cdots & 0 \\ \vdots & \vdots & \ddots & \vdots \\ 0 & 0 & \cdots & \Omega_S \end{pmatrix}$$

$$\Omega_{tot} := \begin{pmatrix} (\Omega_S + \Omega_I^{(1)}) & 0 & \cdots & 0 \\ 0 & (\Omega_S + \Omega_I^{(2)}) & \cdots & 0 \\ \vdots & \vdots & \ddots & \vdots \\ 0 & 0 & \cdots & (\Omega_S + \Omega_I^{(K)}) \end{pmatrix} = \Omega_I + \Omega_S^{tot}$$

$$X_{tot} := \begin{pmatrix} X^{(1)} & 0 & \cdots & 0 \\ 0 & X^{(2)} & \cdots & 0 \\ \vdots & \vdots & \ddots & \vdots \\ 0 & 0 & \cdots & X^{(K)} \end{pmatrix}$$

$$I_K := \begin{pmatrix} I & 0 & \cdots & 0 \\ 0 & I & \cdots & 0 \\ \vdots & \vdots & \ddots & \vdots \\ 0 & 0 & \cdots & I \end{pmatrix}, \text{ where } I \text{ is a } p \times p \text{ identical matrix.}$$

We first introduce the following important Lemma.

**Lemma 1** *For $\varepsilon > 0$ and $a, b \in \mathbb{R}$. If*

$$a, b = \underset{x,y}{argmin} |x| + \varepsilon |y|$$
$$\text{Subject to: } x + y = E \tag{14}$$

*where $E$ is a constant. Then $ab \geq 0$.*

*Proof* If $ab < 0$, let $c = a + b$ and $d = 0$. $cd = 0 \geq 0$ and $c + d = a + b$. Therefore, $|c| + \varepsilon|d| = |a + b| < |a| < |a| + \varepsilon|b|$. This contradicts $a, b$ are the optimal solution of Eq. (14). □

**Corollary 1** *Assume $\widehat{\Omega}_I^{(i)}$ and $\widehat{\Omega}_S$ are the optimal solution of Eq. (7), then* $||(\Omega_S + \Omega_I^{(i)})||_1 = ||\Omega_S||_1 + ||\Omega_I^{(i)}||_1$

*Proof* By Lemma 1, we have that $\widehat{\Omega}_{I,j,k}^{(i)} \widehat{\Omega}_{S,j,k} \geq 0$, if $\widehat{\Omega}_I^{(i)}$ and $\widehat{\Omega}_S$ are the optimal solution of Eq. (7). □

Now we can rewrite our model in Eq. (7) as:

$$\underset{\Omega_{tot}, \Omega_S^{tot}}{argmin} ||\Omega_{tot}||_1 + (\varepsilon - 1)||\Omega_S^{tot}||_1$$
$$\text{Subject to: } ||(\Sigma_{tot}\Omega_{tot} - I_K)||_\infty \leq \lambda_n. \tag{15}$$

This because $||\Omega_{tot}||_1 + (\varepsilon - 1)||\Omega_S^{tot}||_1 = \sum_{i=1}^K ||(\Omega_S + \Omega_I^{(i)})||_1 + K(\varepsilon - 1)||\Omega_S||_1$. By Corollary (1), $||\Omega_{tot}||_1 + (\varepsilon - 1)||\Omega_S^{tot}||_1 = \sum_i ||\Omega_I^{(i)}||_1 + \varepsilon K ||\Omega_S||_1$.

We use $\Sigma_{tot}^0$ to represent the true value of $\Sigma_{tot}$ and $\widehat{\Sigma}_{tot}$ as the estimated. We also use $\Omega_{tot}^0 = (\omega_1^0, \omega_2^0, \ldots, \omega_{Kp}^0)$ to describe the true $\Omega_{tot}$ and $\widehat{\Omega}_{tot}^1 = (\widehat{\omega}_{ij}^1)$ to denote the solution of the optimization problem in Eq. (15). The final solution is denoted as $\widehat{\Omega}_{tot} := (\widehat{\omega}_{ij}) = (\widehat{\omega}_1, \widehat{\omega}_2, \ldots, \widehat{\omega}_{Kp})$ where $\widehat{\omega}_{ij} = \widehat{\omega}_{ji} = \widehat{\omega}_{ij}^1 \text{sign}(\max(|\widehat{\omega}_{ij}^1| - |\widehat{\omega}_{ji}^1|, 0)) + \widehat{\omega}_{ji}^1 \text{sign}(\max(|\widehat{\omega}_{ji}^1| - |\widehat{\omega}_{ij}^1|, 0))$. Furthermore, we denote $\mathbb{E}[X^{(i)}]$ as $(\mu_1^{(i)}, \mu_2^{(i)}, \ldots, \mu_p^{(i)})^T$.

Notice that the constraint of Eq. (15) is not related to $\Omega_S^{tot}$. Therefore $\widehat{\Omega}_{tot}^1$ is also the solution of the following optimization problem.

$$\underset{\Omega_{tot}}{argmin} ||\Omega_{tot}||_1 \quad \text{Subject to: } ||(\Sigma_{tot}\Omega_{tot} - I_K)||_\infty \leq \lambda_n. \tag{16}$$

In the following discussion, we split the analysis into two cases based on two different moment conditions for each $X^{(i)}$.

(C1) *Exponential tails* Suppose there exists a constant $0 < \eta < 0.25$, so that $\frac{\log p}{n_i} \leq \eta$ and

$$\mathbb{E}[e^{t(X_j - \mu_j)^2}] \leq C < \infty, \ \forall \ |t| \leq \eta, \quad \forall j \in \{1, \ldots, p\} \tag{17}$$

where $C$ is a constant.

(C2) *Polynomial-type tails* Suppose that for some $\gamma, c_1 > 0$, $p \leq c_1 n_i^{\gamma}$, and for some $\delta > 0$,

$$\mathbb{E}[|X_j - \mu_j|^{4\gamma+4+\delta}] \leq C < \infty \quad , \quad \forall j \in \{1, \ldots, p\} \tag{18}$$

We begin by assuming that the precision matrix $\Omega$ belongs to the *uniformity class* of matrices,

$$\mathcal{U} := \mathcal{U}(q, s_0(p)) = \left\{ \Omega : \Omega \succ 0, ||\Omega||_1 \leq M, \right.$$

$$\left. \max_{1 \leq i \leq p} \sum_{j=1}^{p} |\omega_{ij}|^q \leq s_0(p) \right\} \tag{19}$$

Here $q$ is a constant and $0 \leq q < 1$. $|| \cdot ||_1$ means the matrix 1-norm. $M$ is a constant representing the upper bound. $\Omega =: (\omega_{ij}) = (\omega_1, \ldots, \omega_p)$. $s_0(p)$ represents the sparsity level of $\Omega$ in the uniformity class. We would like to point out that $s_0(p)$ is related to $p$, but there is no analytic form of relationship between $s_0(p)$ and $p$.

**Lemma 2** *Let* $\{\hat{\Omega}_{tot}^1\}$ *be the solution set of Eq.* (7) *and* $\{\hat{\mathbf{B}}\} := \{(\hat{\mathbf{B}}_I + \hat{\mathbf{B}}_S)\}$, *where*

$$\hat{\mathbf{B}}_I = \begin{pmatrix} \hat{\mathbf{B}}_I^{(1)} & 0 & \cdots & 0 \\ 0 & \hat{\mathbf{B}}_I^{(2)} & \cdots & 0 \\ \vdots & \vdots & \ddots & \vdots \\ 0 & 0 & \cdots & \hat{\mathbf{B}}_I^{(K)} \end{pmatrix}$$

$$\hat{\mathbf{B}}_S^0 = \begin{pmatrix} \hat{\mathbf{B}}_S & 0 & \cdots & 0 \\ 0 & \hat{\mathbf{B}}_S & \cdots & 0 \\ \vdots & \vdots & \ddots & \vdots \\ 0 & 0 & \cdots & \hat{\mathbf{B}}_S \end{pmatrix}$$

*Here* $\hat{\mathbf{B}}_I^{(i)} = (\hat{\beta}_1^{(i)}, \hat{\beta}_2^{(i)}, \ldots, \hat{\beta}_p^{(i)})$ *and* $\hat{\mathbf{B}}_S = (\hat{\beta}_1^s, \hat{\beta}_2^s, \ldots, \hat{\beta}_p^s)$. *For* $j \in \{1, 2, \ldots, p\}$, $(\hat{\beta}_j^{(1)}, \hat{\beta}_j^{(2)}, \ldots, \hat{\beta}_j^{(K)}, \hat{\beta}_j^s)$ *is an optimal solution of Eq.* (9) *when working on the jth column. Therefore,* $\{\hat{\Omega}_{tot}^1\} = \{\hat{\mathbf{B}}\}$.

**Lemma 3** *If each* $\Omega_I^{(i)} + \Omega_S$ *satisfies Condition Eq.* (19), *then* $\Omega_{tot}$ *also satisfies Condition Eq.* (19).

*Proof* use the definition of $\Omega_{tot}$. □

**Corollary 2** $\hat{\Omega}_{tot}$ *satisfies the condition* $\hat{\Omega}_{tot} \succ 0$, *with a high probability.*

*Proof* Lemma 3 indicates that $\Omega_{tot}$ satisfies Condition Eq. (19). □

**Theorem 3** *Suppose that* $\Omega_{tot}^0 \in \mathcal{U}(q, s_0(p))$. *If* $\lambda_n \geq ||\Omega_{tot}^0||_1 (\max_{ij} |\hat{\sigma}_{ij} - \sigma_{ij}^0|)$, *we have that*

$$||\hat{\Omega}_{tot} - \Omega_{tot}^0||_{\infty} \leq 4 ||\Omega_{tot}^0||_1 \lambda_n, \tag{20}$$

$$||\hat{\Omega}_{tot} - \Omega_{tot}^0||_2 \leq C_4 s_0(p) \lambda_n^{1-q}, \tag{21}$$

*and*

$$\frac{1}{p}||\widehat{\Omega}_{tot} - \Omega_{tot}^0||_F \le C_5 s_0(p)\lambda_n^{2-q}, \tag{22}$$

*where constant* $C_4 \le 2(1 + 2^{1-q} + 3^{1-q})(4||\Omega_{tot}^0||_1)^{1-q}$ *and constant* $C_5 \le 4||\Omega_{tot}^0||_1 C_4$.

**Theorem 4** *Suppose that* $\Omega_{tot}^0 \in \mathcal{U}(q, s_0(p))$ *and* $n_{tot} = \sum_{i=1}^K n_i$.
*(a) Assume that condition (C1) holds. Let* $\lambda_n = C_0 M \sqrt{\frac{\log(Kp)}{n_{tot}}}$, *where constant* $C_0 = 2\eta^{-2}(2 + \tau_0 + \eta^{-1}e^2C^2)^2$ *and constant* $\tau_0 > 0$. *Then*

$$||\widehat{\Omega}_{tot} - \Omega_{tot}^0||_2 \le C_1 M^{2-2q} s_0(p) \left(\frac{\log(Kp)}{n_{tot}}\right)^{(1-q)/2} \tag{23}$$

$$||\widehat{\Omega}_{tot} - \Omega_{tot}^0||_\infty \le 4C_0 M^2 \sqrt{\frac{\log(Kp)}{n_{tot}}} \tag{24}$$

$$\frac{1}{p}||\widehat{\Omega}_{tot} - \Omega_{tot}^0||_F^2 \le 4C_1 M^{4-2q} s_0(p) \left(\frac{\log(Kp)}{n_{tot}}\right)^{1-q/2} \tag{25}$$

*with a probability greater than* $1 - 4p^{-\tau_0}$, *where constant* $C_1 \le 2(1 + 2^{1-q} + 3^{1-q})4^{1-q}C_0^{1-q}$.

*(b) Assume that condition (C2) holds. Let* $\lambda_n = C_2 \sqrt{\frac{\log(Kp)}{n_{tot}}}$, *where constant* $C_2 = \sqrt{(5 + \tau_0)(\theta_0 + 1)}$ *and*

$$\theta_0 = \max_{i,j,k} \hat{\Sigma}_{j,k}^{(i)}. \tag{26}$$

*Then*

$$||\hat{\Omega}_{tot}^0 - \Omega_{tot}^0||_2 \le C_3 M^{2-2q} s_0(p) \left(\frac{\log(Kp)}{n_{tot}}\right)^{(1-q)/2} \tag{27}$$

$$||\widehat{\Omega}_{tot} - \Omega_{tot}^0||_\infty \le 4C_2 M^2 \sqrt{\frac{\log(Kp)}{n_{tot}}} \tag{28}$$

$$\frac{1}{p}||\widehat{\Omega}_{tot} - \Omega_{tot}^0||_F^2 \le 4C_3 M^{4-2q} s_0(p) \left(\frac{\log(Kp)}{n_{tot}}\right)^{(1-q)/2} \tag{29}$$

*with a probability greater than* $1 - O(n_{tot}^{-\delta/8} + p^{-\tau_0/2})$, *where constant* $C_3 \le 2(1 + 2^{1-q} + 3^{1-q})4^{1-q}C_2^{1-q}$.

Through Eq. (23)–(29), we theoretically prove that we can achieve a good estimation of target dependency graphs with the convergence rate $O(\log(Kp)/n_{tot})$. Based on CLIME (Cai et al. 2011), the convergence rate of single-task sGGM is $O(\log p/n_i)$. Here $n_i$ represents the number of samples of $i$th task. Assuming $n_i = \frac{n_{tot}}{K}$, the convergence rate of single sGGM is $O(K \log p/n_{tot})$. Clearly, since $K \log p > \log(Kp)$, the convergence rate of SIMULE is better than single-task sGGM. This provides theoretical proofs for the benefit of multi-tasking sGGM. Neither of these theoretical results have been investigated by the previous studies.

All proofs of above theorems are provided in Sect. 7.

## 5.2 Theoretical analysis for NSIMULE estimator

In this subsection, we investigate the theoretical properties of NSIMULE estimator and prove that its convergence rate is the same as the SIMULE.

**Theorem 5** *Based on* Liu et al. (2009), *Suppose we use the estimated Kendall's tau correlation matrix* $\widehat{\mathbf{S}}$ *to replace* $\widehat{\Sigma}$ *in the parametric GLasso estimator. Then under the same conditions on* $\widehat{\Sigma}$ *(that ensure the consistency of the estimator under the Gaussian model), the nonparanormal estimator achieves the same (parametric) rate of convergence as GLasso estimator for both the precision matrix estimation and the graph structure recovery.*

**Theorem 6** *We use the estimated Kendall's tau correlation matrix* $\hat{\mathbf{S}}^{(i)}$ *in the Nonparanormal SIMULE estimator. Then under the same conditions on* $\hat{\Sigma}^{(i)}$ *(that ensure the consistency of the estimator under the Gaussian model), the nonparanormal SIMULE achieves the same rate of convergence as SIMULE* $(O(\sqrt{\log(Kp)/n_{tot}}))$ *for both the graph recovery and precision matrix estimation.*

*Proof* We can directly apply Theorem 5 (the main theorem of Liu et al. 2009). Our multi-task nonparanormal graph estimation will not change the convergence rate of SIMULE. □

### 5.3 Potential non-identifiability issue

Linear programming is not strongly convex. Therefore, there may be multiple ideal solutions in the SIMULE formulation of Eq. (7) (i.e., identifiability problem). In fact, the CLIME (Cai et al. 2011) estimator may also have multiple optimal solutions. Cai et al. (2011) have proved all such solutions converge to the true one at an optimal convergence rate. Similarly, in Sect. 5.1, we have proved that SIMULE formulation in Eq. (7) may result in multiple optimal solutions $\{\hat{\Omega}_{tot}\}$. Each of these solutions $\hat{\Omega}_{tot}$ converges to the true solution with an optimal convergence rate. We present Theorem 7 showing that for each optimal solution $\hat{\Omega}_{tot}$, when $\varepsilon \neq 1$, we can obtain unique estimation of $\Omega_S$ and $\{\Omega_I^{(i)}|i=1,\ldots,K\}$.

**Theorem 7** *When we pick* $\varepsilon > 0$ *and* $\varepsilon \neq 1$, *for each optimal solution* $\hat{\Omega}_{tot}$ *from Eq. (7), there exist unique* $\Omega_S$ *and* $\{\Omega_I^{(i)}|i=1,\ldots,K\}$ *satisfying Eq. (5).*

We provide the proof of Theorem 7 in Sect. 7.

In practice, we need to decide $\varepsilon$ according to the application for which SIMULE is used. For example, for the genome-related biomedical data, we can normally assume that the shared subgraph is more dense than individual interactions of each context.[4] Therefore we pick $\varepsilon < 1$ to reflect this assumption in our experiments in Sect. 6.

## 6 Experiments

In this section, we use four simulated datasets and two real-world bio-datasets to evaluate the proposed estimators.

### 6.1 Experimental settings

#### 6.1.1 Baselines

We compare SIMULE, SIMULE-I and NSIMULE with the following baselines: (1) Three different multi-sGGM estimators including JGL-fused, JGL-group (Danaher et al. 2013), and SIMONE (Chiquet et al. 2011) (with the penalty functions described in Table 1); (2) The

---

[4] This assumes more interactions are preserved across cell contexts, i.e., partly due to concerns of system or evolutionary stability.

**Table 3** AUC and partial AUC on simulated Gaussian datasets from Model 1

| | AUC | AUC-individual | AUC-shared | AUC-FPR ≤ 20% | AUC-FPR ≤ 5% | AUC-$p = 200$ |
|---|---|---|---|---|---|---|
| *Gaussian-model1*-[$K = 3$]-[$p = 100$] | | | | | | |
| NSIMULE | **0.9872** | **0.8408** | 0.8964 | **0.1599** | **0.0188** | **0.9959** |
| SIMULE | **0.9844** | **0.8379** | 0.8788 | **0.1587** | **0.0179** | **0.9945** |
| JGL-fused | 0.6843 | 0.5666 | **0.9817** | 0.0989 | 0.0094 | 0.6745 |
| JGL-group | 0.5162 | 0.4988 | 0.5759 | 0.0908 | 0.0174 | 0.5122 |
| SIMONE | 0.7748 | 0.5124 | **0.9321** | 0.0992 | 0.0171 | 0.5488 |
| CLIME | 0.6509 | 0.5197 | 0.7795 | 0.0439 | 0.0001 | 0.5422 |
| NCLIME | 0.5400 | 0.4999 | 0.8224 | 0.0434 | 0.0001 | 0.5216 |
| SIMONE-I | 0.8041 | 0.6740 | **0.9681** | 0.1030 | 0.0177 | 0.6122 |
| SIMULE-I | **0.9979** | **0.8594** | 0.9249 | **0.1604** | **0.0183** | **0.9984** |

For each column, we make the font bold for the numbers of the top three performed methods

**Table 4** AUC and partial AUC on simulated nonparanormal datasets from Model 1

| | AUC | AUC-individual | AUC-shared | AUC-FPR ≤ 20% | AUC-FPR ≤ 5% | AUC-$p = 200$ |
|---|---|---|---|---|---|---|
| *Nonparanormal-model1*-[$K = 3$]-[$p = 100$] | | | | | | |
| NSIMULE | **0.8172** | **0.8408** | 0.8964 | **0.1599** | **0.0188** | **0.8325** |
| SIMULE | **0.7322** | **0.8165** | 0.8788 | **0.1567** | **0.0173** | **0.7745** |
| JGL-fused | 0.6942 | 0.6362 | **0.9817** | 0.0993 | 0.0124 | 0.7308 |
| JGL-fused-nonparanormal | 0.6978 | **0.6374** | **0.9896** | 0.1012 | **0.0137** | **0.7343** |
| JGL-group | 0.5181 | 0.4942 | 0.8050 | 0.0487 | 0.0064 | 0.5416 |
| JGL-group-nonparanormal | 0.6500 | 0.5563 | 0.8614 | 0.0868 | 0.0113 | 0.5498 |
| SIMONE | 0.7198 | 0.5061 | 0.9321 | 0.1080 | 0.0102 | 0.5671 |
| SIMONE-nonparanormal | **0.7271** | 0.5072 | **0.9327** | **0.1146** | 0.0135 | 0.5766 |
| CLIME | 0.5803 | 0.5108 | 0.7745 | 0.0427 | 0.0001 | 0.5127 |
| NCLIME | 0.5298 | 0.4935 | 0.8224 | 0.0398 | 0.0001 | 0.4915 |

For each column, we make the font bold for the numbers of the top three performed methods single-task CLIME baseline (i.e. each task uses CLIME independently); (3) The single-task nonparanormal CLIME (NCLIME) baseline (i.e., each task uses NCLIME independently). (4) The nonparanormal extension of JGL-fused, JGL-group, and SIMONE.[5] (5) SIMONE-I ("intertwined Lasso" in SIMONE package) is added as a baseline for comparison with SIMULE-I.[6]

### 6.1.2 Metric

The edge-level false positive rate (FPR) and true positive rate (TPR) are used to measure the difference between the true graphs and the predicted graphs. By repeating each graph

---

[5] We extend JGL and SIMONE on nonparanormal distributions by replacing the codes for sample covariance matrix into Kendall's tau correlation matrix in their R implementations.

[6] It is possible to combine nonparanormal and intertwined strategies to extend SIMULE. We leave this as a future work.

**Table 5**  AUC and partial AUC on simulated Gaussian datasets from Model 2

|  | AUC | AUC-individual | AUC-shared | AUC-FPR ≤ 20% | AUC-FPR ≤ 5% | AUC-$p = 200$ |
|---|---|---|---|---|---|---|
| *Gaussian-model2-[K = 2]* | | | | | | |
| NSIMULE | **0.9997** | **0.8095** | 0.9727 | **0.1997** | **0.0497** | **1.0000** |
| SIMULE | 0.9996 | **0.8391** | 0.9697 | **0.1997** | **0.0497** | **0.9998** |
| JGL-fused | 0.9991 | 0.4893 | **0.9983** | 0.1991 | 0.0491 | 0.9993 |
| JGL-group | **0.9999** | 0.5000 | 0.7715 | 0.1999 | **0.0499** | 0.9866 |
| SIMONE | 0.9989 | 0.7632 | 0.9982 | 0.1989 | 0.0489 | 0.9990 |
| CLIME | 0.5077 | 0.3948 | 0.7517 | 0.0404 | 0.0025 | 0.5037 |
| NCLIME | 0.4995 | 0.4043 | 0.7614 | 0.0402 | 0.0025 | 0.5022 |
| SIMONE-I | 0.9995 | 0.7692 | **0.9986** | 0.1995 | 0.0499 | 0.9951 |
| SIMULE-I | **0.9997** | **0.8704** | **0.9997** | 0.1999 | 0.0499 | **1.0000** |

For each column, we make the font bold for the numbers of the top three performed methods

**Table 6**  AUC and partial AUC on simulated nonparanormal datasets from Model 2

|  | AUC | AUC-individual | AUC-shared | AUC-FPR ≤ 20% | AUC-FPR ≤ 5% | AUC-$p = 200$ |
|---|---|---|---|---|---|---|
| *Nonparanormal-model2-[K = 2]* | | | | | | |
| NSIMULE | **0.9993** | **0.8095** | 0.9727 | **0.1993** | **0.0493** | **1.0000** |
| SIMULE | **0.9993** | **0.8453** | 0.9609 | 0.1929 | 0.0419 | **0.9996** |
| JGL-fused | 0.9984 | 0.5117 | **0.9984** | 0.1984 | 0.0424 | 0.9996 |
| JGL-fused-nonparanormal | 0.9990 | 0.5641 | **0.9998** | 0.1986 | 0.0425 | 0.9996 |
| JGL-group | 0.9784 | 0.6791 | 0.9151 | 0.1967 | 0.0464 | **1.0000** |
| JGL-group-nonparanormal | 0.9899 | 0.6948 | 0.9391 | 0.1969 | 0.0465 | **1.0000** |
| SIMONE | 0.9991 | 0.7529 | 0.9911 | 0.1990 | **0.0491** | 0.9983 |
| SIMONE-nonparanormal | **0.9992** | **0.7960** | **0.9948** | **0.1993** | 0.0492 | 0.9986 |
| CLIME | 0.4985 | 0.3740 | 0.7517 | 0.0238 | 0.0001 | 0.5041 |
| NCLIME | 0.4994 | 0.4042 | 0.7614 | 0.0203 | 0.0001 | 0.5022 |

For each column, we make the font bold for the numbers of the top three performed methods inference process 10 times, we obtain average metrics (also being averaged over $K$ tasks) for each method we test. Here, FPR $= \frac{FP}{FP + TN}$ and TPR $= \frac{TP}{TP + FN}$. For a predicted graph, the TP (true positive) and TN (true negative) mean the number of true nonzero entries and true zero entries respectively. Each FPR vs. TPR curve shows the multi-point performance of a method over a range of its regularization parameter. In Tables 3, 4, 5 and 6 we compare estimators using the area under a FPR–TPR curve (AUC) and the partial area (partial AUC) under a FPR–TPR curve within the range of $FPR < q$ (e.g., $q = 20\%$ or $q = 5\%$). For instance, when we choose $q = 5\%$ we care about the partial AUC scores of those regions with very small FPRs. Higher AUC or partial AUC scores indicate that a method has achieved better results overall.

### 6.1.3 Selection of hyper-parameter $\lambda_n$

Recent research studies from Negahban et al. (2009) and Yang et al. (2014) conclude that the regularization parameter $\lambda$ of a single-task sGGM (e.g., with $n_i$ samples) should satisfy $\lambda \propto \sqrt{\frac{\log p}{n_i}}$. Combining this conclusion with our theoretical analysis in Sect. 5, we choose $\lambda_n = \alpha \sqrt{\frac{\log(Kp)}{n_{tot}}}$ where $\alpha$ for SIMULE or NSIMULE is a hyper-parameter to tune. In our experiments, $\alpha$ is varied over a range of $\{0.05 \times i \mid i \in \{1, 2, 3, \ldots, 30\}\}$. The Bayesian information criterion (BIC) is used for situations requiring to select a specific value of hyper-parameters.

Besides, we also need to tune the hyper-parameters of the baseline methods to obtain their FPR–TPR curves. If only one hyper-parameter needs tuning, we follow the same strategy as SIMULE. For those baselines (JGL-fused and JGL-group) having two hyper-parameters, when given a certain $\lambda_1$ (the same as $\lambda_n$), we use BIC criteria to select its best $\lambda_2$ from a range of $\{0.05 \times i \mid i \in \{1, 2, 3, \ldots, 20\}\}$.

### 6.1.4 Selection of hyper-parameter $\varepsilon$

$\varepsilon$ reflects the difference of sparsity in the shared subgraph versus the context-specific subgraphs. Section 5.3 has discussed our choice of $\varepsilon$ on two real-world datasets. Similarly for the simulated experiments, we select $\varepsilon$ from a range of $\{0.1 \times i \mid i \in \{1, 2, \ldots, 9\}\}$.

### 6.2 Simulated Gaussian datasets

Using the following two graph models, we first generate two synthetic multivariate Gaussian datasets, in which each model includes $K$ tasks of data samples.

- *Model 1* Coming from Rothman et al. (2008), this model assumes $\Omega^{(i)} = \mathbf{B}_I^{(i)} + \mathbf{B}_S + \delta^{(i)} I$, where each off-diagonal entry in $\mathbf{B}_I^{(i)}$ is generated independently and equal to 0.5 with probability $0.05i$ and 0 with probability $1 - 0.05i$. The shared part $\mathbf{B}_S$ is generated independently and equal to 0.5 with probability 0.1 and 0 with probability 0.9. $\delta^{(i)}$ is selected large enough to guarantee the positive definiteness of precision matrix. A clear shared structure $\mathbf{B}_S$ exists among multiple graphs. We choose $K \in \{2, 3, 4, 5, 6\}$ for this case.
- *Model 2* This model uses two special-structure graphs, i.e., a grid graph and a ring graph. The first task uses a grid graph and the second uses a ring graph. These two special networks are popular in many real-world applications . For instance, certain biological pathways can be represented as rings. A clear shared structure exists between these two graphs. Clearly, $K = 2$ for this case.

We choose $p = 100$, i.e., the dimension of feature variables is 100. For each dataset, 500 data samples are generated randomly. Using either Model 1 or Model 2, we generate $K$ blocks of data samples with the $i$th block following $N(0, (\Omega^{(i)})^{-1})$. In detail, for $i$th task, we generate 500 simulated data samples following multivariate Gaussian Distribution with mean 0 and covariance matrix $(\Omega^{(i)})^{-1}$. We use the multivariate distribution method from stochastic simulation (Ripley 2009) to sample the simulated data blocks. In our implementation, we directly use the R function "*mvrnorm*" in *MASS* package.

In summary, we first simulate precision matrices by Model 1 or Model 2. We then use multivariate distribution method (Ripley 2009) to sample multivariate Gaussian Distributed data blocks with mean 0 and covariance matrix $(\Omega^{(i)})^{-1}$. This stochastic procedure will generate simulated data blocks with the decomposition in Eq. (5). Then we apply SIMULE, NSIMULE and baseline models on these datasets to obtain the estimated dependency networks.

### 6.3 Simulated nonparanormal datasets

Using the same graphs of Model 1 and Model 2, we simulate two more sets of data samples following the nonparanormal distributions in $K$ different tasks. We pick $K = 3$ for Model 1 and $K = 2$ for Model 2. Starting from $N(0, (\Phi^{(i)})^{-1})$ and transforming with a monotone function $x :\to sign(\mathbf{x})|\mathbf{x}|^{\frac{1}{2}}$, data sample is generated as a random vector $Z$, where $sign(Z)Z^2 = (sign(\mathbf{z}_1)\mathbf{z}_1^2, \ldots, sign(\mathbf{z}_p)\mathbf{z}_p^2) \sim N(0, (\Phi^{(i)})^{-1})$. Thus $Z$ follows a nonparanormal distribution. Using different monotone functions for generating simulated data will not change the resulting correlation matrix $\mathbf{S}$ since we use the rank-based nonparametric estimator estimating the correlation matrix.

### 6.4 Experimental results on synthetic datasets

Using Tables 3, 4, 5 and 6, we compare all methods using (a) the area under a FPR–TPR curve (AUC), (b) the area under a FPR–TPR curve within the range of $FPR < 0.2$ (AUC-20%), (c) the area under a FPR–TPR curve within the range of $FPR < 0.05$ (AUC-5%), (d) the AUC scores when evaluating only on the shared part (AUC-shared), (e) the AUC scores when evaluating only on the individual parts of multiple graphs (AUC-individual), and (f) AUC scores when changing $p$ (AUC-$p = 200$). Tables 3 and 4 include results on datasets from graph Model 1. Tables 5 and 6 are about Model 2. Tables 3 and 5 show comparisons on the simulated Gaussian datasets. Tables 4 and 6 are about results on the simulated nonparanormal datasets. Besides, Tables 4 and 6 also provide the results of the nonparanormal implementations of three baselines[7]—JGL-fused, JGL-group and SIMONE. In summary, we can conclude from these tables that (1) SIMULE and NSIMULE methods obtain better AUC and partial AUC than three multi-sGGM competitors, nonparanormal extensions of these three competitors, and two single-sUGM baselines across Model 1 and Model 2, and across two different values of $p$. (2) NSIMULE produces better scores than SIMULE on nonparanormal data across both graph models (more apparent on Model 1). Similarly, nonparanormal extensions of three multi-sGGM baselines outperform their Gaussian-versions on the nonparanormal datasets. (3) All multi-task estimators perform better than single-CLIME and single-NCLIME estimators. (4) On datasets from Model 2, the score differences among multi-task estimators are not as apparent as the case of Model 1. (5) SIMULE and NSIMULE achieve better estimation of the individual sub-graphs than other baselines. (6) The columns of "AUC-shared" show that JGL-fused and SIMONE estimate the shared parts better. This is as expected, since their penalty functions have enforced the shared similarity among graphs. (7) Interestingly, in Tables 3 and 5 NSIMULE achieves similar or even better AUC scores on Gaussian datasets. This is actually also as expected, since under a high-dimensional setting the Kendalls tau correlation matrix used in NSIMULE has been proved to provide consistent estimation of correlation matrix with an optimal convergence rate (Liu et al. 2009). Differently, sample covariance matrix used by SIMULE is inconsistent for high-dim cases (Fan et al. 2014). As a future step, large covariance-estimators like POET (Fan et al. 2014) will be used to replace sample-covariance in SIMULE. (8) Related to the previous point, SIMULE-I and SIMONE-I have revised context-specific sample covariance matrices with a bias towards a global empirical covariance matrix over all $K$ tasks. The last two rows in Tables 3 and 5 clearly show that such a simple strategy on covariance estimation has provided significant performance improvement over both SIMULE and SIMONE.

---

[7] We revise the R packages of three baselines to extend them to nonparanormal distribution. This is the provide a more fair comparison of these baselines versus NSIMULE.

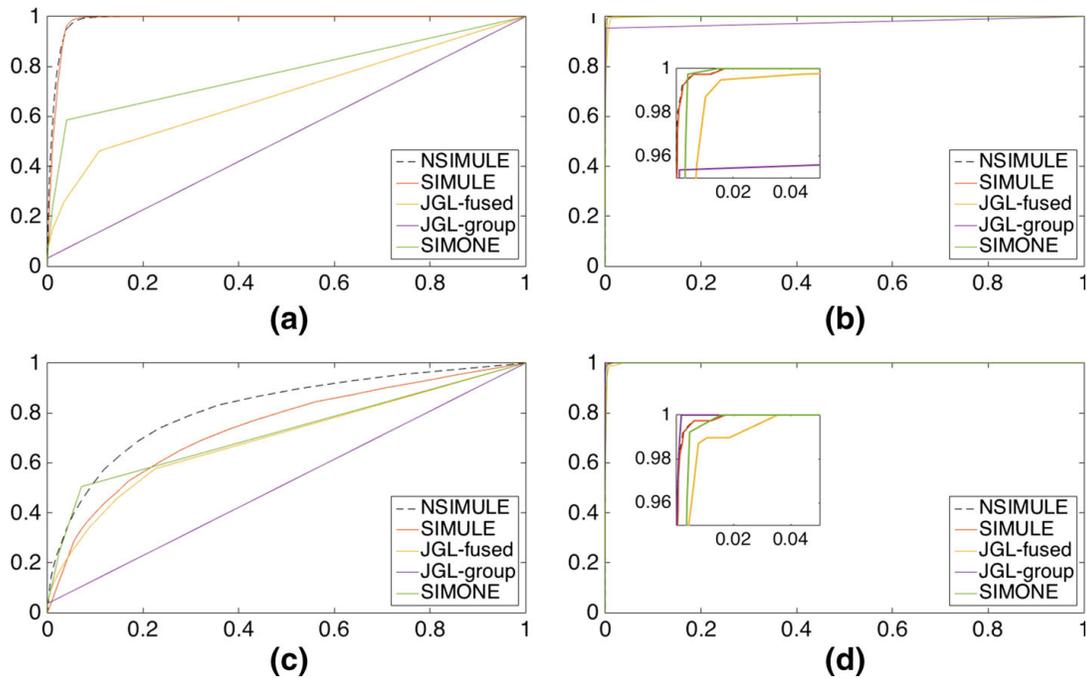

**Fig. 3** The FPR–TPR curve graph for different methods on four simulated datasets. The *upper two* are for Gaussian simulated datasets. The lower two are for nonparanormal datasets. The *upper left* and *lower left graphs* are generated using graph Model 1. The *upper right* and *lower right* are generated using Model 2. Each *curve* is generated by varying the λ parameter(s). We can see that curves from SIMULE and NSIMULE are above all baseline multi-sGGM methods (more apparent on two datasets generated by Model 1)

Figure 3 shows more detailed FPR–TPR comparisons among the joint sGGM estimators. The sub-figure (a) "Gaussian-Model1" clearly shows that our methods obtain better curves than three multi-sGGM baselines. On the sub-figure (b) "Gaussian-Model2", the differences among multi-sGGM estimators are not as apparent as "Gaussian-Model1". Figure 3c, d (in the second row) show FPR–TPR curves from two nonparanormal datasets. Overall, the observations from comparing these curves are consistent with those obtained from the four tables (Tables 3, 4, 5, 6).

Since the difference among multi-task estimators is not apparent on the datasets from Model 2, we further visualize the derived graph structures from all joint sGGM estimators using a black-white color-map visualization (i.e., black for nonzero entries and white for zero entries of precision matrix) in Fig. 4. The left two columns of Fig. 4 present all color maps we have obtained on "Gaussian-Model 2". The right two columns show all color maps from "nonparanormal-Model 2". The first row represents the original grid graph and ring graph from Model 2. Since these two graph structures are clearly related, we can observe a diagonal band shared by the two color-maps. "Graph 1" grid graph has two more off-diagonal narrow bands while "Graph 2" ring graph does not. The color maps of predicted graphs with proper tuning parameters from the NSIMULE and SIMULE method s are presented as the second row and the third row. Three other multi-sGGM baselines are shown on the fourth, fifth and the sixth rows. Instead of pushing two graphs to be too similar, SIMULE and NSIMULE methods achieve a better recovery of both the ring graph and the grid graph. In contrast, three multi-task baseline methods all obtain an incorrect graph on the second task (ring graph) by pushing it to be too similar to the first task (grid graph). The single task baselines (CLIME as the 8th row and NCLIME as the 7th row) obtain two much worse graphs because they do not model the relationship among the two graphs at all.

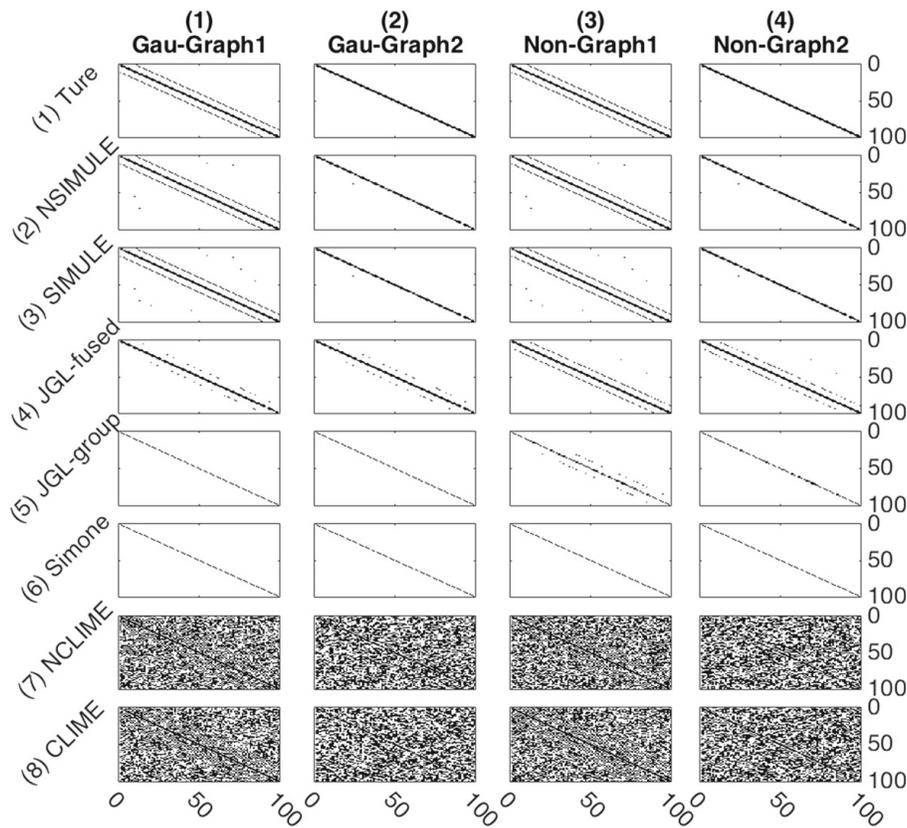

**Fig. 4** On two simulated datasets generated from Model 2, the four estimated conditional-dependency graphs from all methods are compared using black-white color-maps. In total we visualize two true graph (grid and ring) and fourteen predicted dependency networks (by SIMULE, NSIMULE, three multi-sGGM baselines, CLIME and NCLIME). The *left two columns* are derived from the simulated Gaussian Model 2 dataset. The *right two columns* are derived from the simulated nonparanormal Model 2 dataset. We can conclude that three multi-sGGM baseline methods obtain incorrect recoveries since they push two tasks to be similar through penalty functions. In contrast, SIMULE and NSIMULE [row (2), (3)] achieve better edge predictions by modeling both joint and task-specific parameters

One important hyper-parameter we need to pick for SIMULE and NSIMULE is $\varepsilon$ in Eq. (7). This hyper-parameter reflects the sparsity level of the shared subgraph against the context-specific parts. The left two sub-figures (a) and (c) of Fig. 5 show the changes of sparsity level in both individual and shared subgraphs across multiple values of $\varepsilon$ (by running SIMULE on the Gaussian-Model 1 case). Figure 5a is for $K = 3$ and Fig. 5c is for $K = 6$. We can see that when $\varepsilon$ increases, the sparsity level of shared portion decreases while the averaging sparsity of individual parts increases across both cases of $K$. This matches our analysis in Sect. 5.3. In real applications, $\varepsilon$ indicates the differences of sparsity constraints we assume on shared and individual parts. It should be chosen according to the domain knowledge of a specific application.

In addition, Fig. 5e tries to investigate whether the changes of $\varepsilon$ influence the performance (AUC) of (N)SIMULE. On datasets from Model 1, Fig. 5d shows that AUC scores of both SIMULE and NSIMULE exhibit very small variations across a large range of changing $\varepsilon$.

Section 2.5 presents a parallel variation of SIMULE. On the simulated data of Gaussian-Model 1, we compare the training speed of original-SIMULE versus parallel-SIMULE using Fig. 5b (the right upper sub-figure). For the parallel-SIMULE, we run SIMULE by paralleling "column per core" using 63 cores on a 64-core machine. The baselines, including the original-

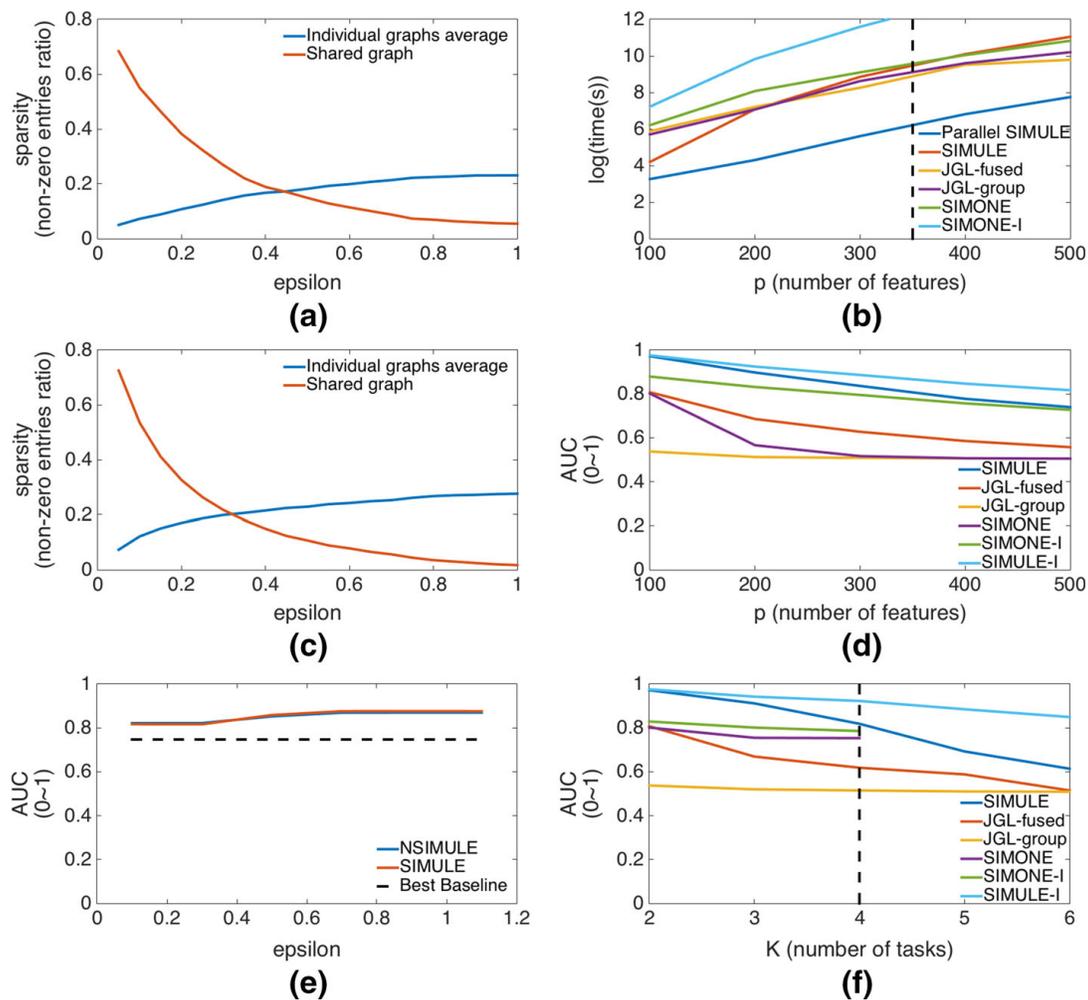

**Fig. 5** **a**, **c** Changes of sparsity level of the individual and the shared subgraphs when varying the hyperparameter $\varepsilon$. **e** Changes of AUC when varying the hyperparameter $\varepsilon$. The results in (**a**), (**c**), (**e**) are obtained from running SIMULE on Gaussian-Model 1 case. The *upper right sub-figure* (**b**) uses the training time (in log-seconds) versus the varying $p$, to compare the "column-per-core" parallel implementation of SIMULE with the original SIMULE implementation, JGL-fused, JGL-group, SIMONE and SIMONE-I. *Right sub-figures* (**d**, **f**) show AUC scores from SIMULE, three multi-sGGM baselines, SIMONE-I and SIMULE-I against varying $p$ and against varying $K$

SIMULE, JGL-fused, JGL-group, SIMONE and SIMONE-I, are run on one restricted core on the same machine.[8] Figure 5b provides the computational speed (training time in log-seconds) across values of dimension $p$. It clearly shows that the parallel-SIMULE runs much faster than the single-core SIMULE implementation and other baselines.[9]

Furthermore, Fig. 5d, f provide AUC scores of SIMULE, three multi-SGGM baselines, SIMONE-I and SIMULE-I against the varying $p$ and against the with varying $K$. SIMULE-I provides a consistent performance improvement over SIMULE. SIMULE outperforms three multi-sGGM baselines across multiple values of $p$ and $K$. SIMULE-I outperforms SIMONE-I and runs smoothly for larger values of $K$. Unfortunately Fig. 5f can not provide the AUC scores of SIMONE-I for $K = 5$ and $K = 6$, because the SIMONE package could not converge for some $\lambda$ values under these two cases of $K$.

---

[8] The multi-core setting we use for this time experiment is to reflect the distributed parallel nature of SIMULE. We leave the topic of using the multi-threading to improve SIMULE as future research.

[9] When $p \geq 400$, SIMONE-I takes more than 5 days to train. That's why the no data points are shown in Fig. 5b for such cases.

### 6.5 Experiment results on real application I: identifying gene interaction using gene expression data across two cell contexts

Next, we apply SIMULE and the baselines on one real-world biomedical data: gene expression profiles describing many human samples across multiple cancer types (aggregated by McCall et al. 2011). Recently advancements in genome-wide monitoring have resulted in enormous amounts of data across most of the common cell contexts, like multiple common cancer types (The Cancer Genome Atlas Research Network 2011). Complex diseases such as cancer are the result of multiple genetic and epigenetic factors. Thus, recent research has shifted towards the identification of multiple genes/proteins that interact directly or indirectly in contributing to certain disease(s). Structure learning of UGMs on such heterogeneous datasets can uncover statistical dependencies among genes and understand how such dependencies vary from normal to abnormal or across different diseases. These structural variations are highly likely to be contributing markers that influence or cause the diseases.

Two major cell contexts are selected from the human expression dataset provided by McCall et al. (2011): leukemia cells (including 895 sample s) and normal blood cells (including 227 samples). Then we choose the top 1000 features from the total 12,704 features (ranked by variance) and apply SIMULE[10], JGL-fused, JGL-group, SIMONE, single-CLIME, SIMONE-I and SIMULE-I on this two-task dataset. The derived dependency graphs are compared by using the number of predicted edges being validated by three major existing protein/gene interaction databases (Prasad et al. 2009; Orchard et al. 2013; Stark et al. 2006).[11] The numbers of matches between interactions in databases and those edges predicted by each method have been shown as a bar graph in the left subfigure of Fig. 6. This bar graph clearly shows that SIMULE consistently outperforms three multi-sGGM baselines and CLIME on both individual and shared interactions from both cell contexts. Interestingly, SIMULE-I outperforms SIMONE-I and SIMULE. SIMONE-I achieves a better recovery of known edges than JGL-group, JGL-fused, SIMONE and CLIME. This leaves us to believe that exploring more choices of covariance matrix estimation to extend SIMULE is promising.

### 6.6 Experiment results on real application II : identifying collaborations among TFs across multiple cell types

In molecular biology, the regulatory proteins that interact with one another to control gene transcription are known as transcription factors (TFs). TF proteins typically perform major cell regulatory functions (e.g., binding on DNA) by working together with other TFs. The collaboration patterns (e.g., conditional independence) among TFs normally vary across different cell contexts (e.g., cell lines). Meanwhile, a certain portion of the TF interactions are preserved across contexts. Understanding the collaboration networks among TFs is the key to understanding cell development, including defects, which lead to different diseases. The ChIP-Seq datasets recently made available by the ENCODE project (ENCODE Project Consortium 2011) provide simultaneous binding measurements of TFs to thousands of gene

---

[10] NSIMULE was tried as well and has achieved the same validation result as SIMULE.

[11] We would like to point out that the interactions SIMULE finds represent statistical dependencies between genes that vary across multiple cell types. There exist many possibilities for such interactions, including like physical protein-protein interactions, regulatory gene pairs or signaling relationships. Therefore, we combine multiple existing databases for a joint validation.

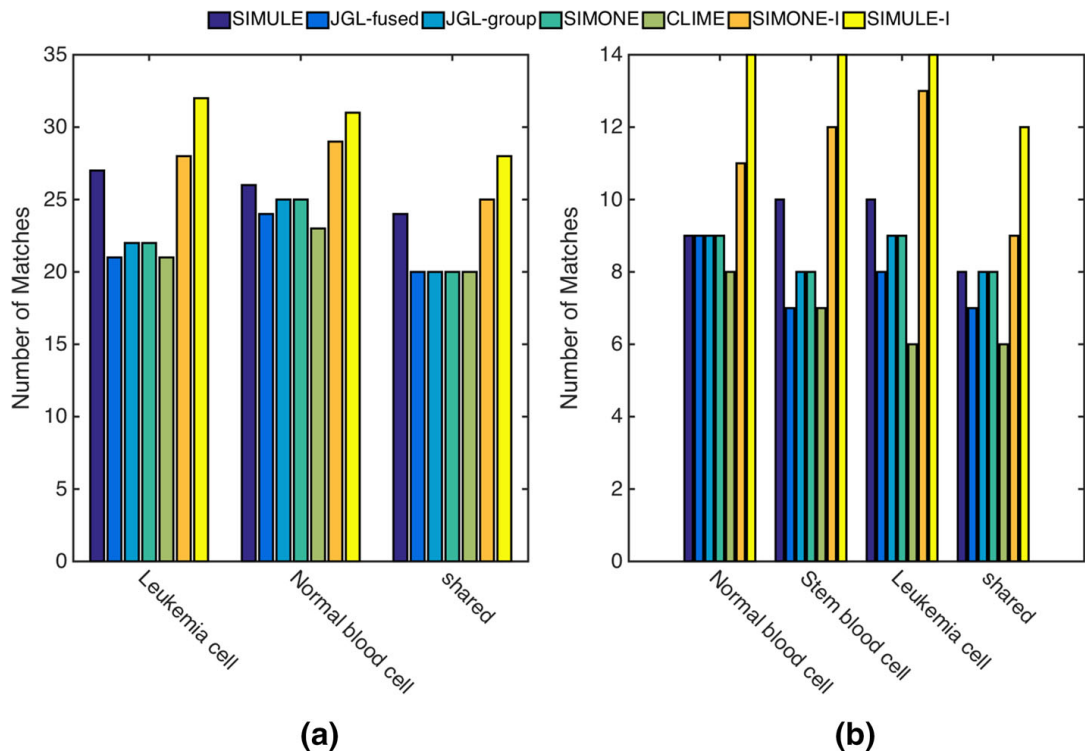

**(a)** **(b)**

**Fig. 6** Validating the recovered molecule dependencies using existing bio-databases (Prasad et al. 2009; Orchard et al. 2013; Stark et al. 2006) of experimentally validated known protein-protein interactions (PPI) or gene interactions in humans. The *left sub-figure* (**a**) is obtained from a multi-cell gene expression data. The *right sub-figure* (**b**) is obtained from a multi-cell TF-binding data. The "shared" bars from the baseline methods are obtained using the intersection among the predicted graphs. The number of matches among predicted edges and known interactions is shown as *bar lines*

targets. These measurements provide a "snapshot" of TF binding events across many cell contexts. The task of uncovering the functional dependencies among TFs connects to the task of discovering the statistical dependencies among TFs from their ChIP-Seq measurements.

Recently, two relevant papers (Cheng et al. 2011; Min et al. 2014) have discussed methods to infer co-association networks among TFs using ChIP-Seq data. Their approaches differ from ours as both projects has targeted a single cell type at a time. We select ChIP-Seq data for 27 TFs that are covered by ENCODE (ENCODE Project Consortium 2012) across three major human cell lines including (1) H1-hESC (embryonic stem cells : primary tissue), (2) GM12878 (B-lymphocyte:normal tissue) and (3) K562 (leukemia:cancerous tissue).

We apply SIMULE, SIMULE-I, JGL-group, JGL-fused, SIMONE and SIMONE-I to this multi-cell TF ChIP-Seq dataset. Comparisons of different methods are performed using three major existing protein interaction databases (Prasad et al. 2009; Orchard et al. 2013; Stark et al. 2006). The numbers of matches between TF-TF interactions in databases and those predicted by each method have been plotted as a bar graph shown in Fig. 6b (the right sub-figure). The graph shows that SIMULE consistently outperforms JGL-fused, JGL-group and SIMONE on both individual and shared interactions from all three cell types. SIMULE-I performs better than SIMONE-I and SIMULE. We further evaluated the resulting TF interactions using the popular "functional enrichment" analysis with DAVID (Da Wei Huang and Lempicki 2008) and found that SIMULE and SIMULE-I can reveal known functional sets and potentially novel interactions that drive leukemia. This leads us to believe that our approach can be used in a wider range of applications as well.

Many domain-specific studies have applied sGGM on real-world datasets, especially those from molecular biology or brain science. For example, Ma et al. (2007) estimates gene networks using miroarray data in the model species Arabidopsis. Krumsiek et al. (2011) explores GGM to reconstruct pathway reactions from high-throughput metabolomics data. Within brain science, a few studies (Ng et al. 2013; Huang et al. 2010; Monti et al. 2015; Sun et al. 2009) have tried to learn brain connectivity of Alzheimer's disease through sparse inverse covariance estimation. Due to space limit, we omit detailed inclusions of these studies.

# 7 Proof of Theorems

In this section, $I(S)$ denotes the indicator function of the set $S$.

## 7.1 Lemma 2

*Proof* In the following, we just show the case of $1 \leq j \leq p$. When $j > p$, by replacing $e_{j \mod p}$ in the following proofs, it will still hold. $||\Sigma_{tot} \widehat{\Omega}_{tot}^1 - I||_\infty \leq \lambda_n$ is equivalent to $||\Sigma_{tot} \widehat{\omega}_j^1 - e_j||_\infty \leq \lambda_n$ for $1 \leq j \leq p$.

Thus we have

$$\left|\left|\widehat{\omega}_j^1\right|\right|_1 \geq ||\hat{\beta}_j||_1 \ \ \forall 1 \leq j \leq p. \tag{30}$$

Also $||\Sigma_{tot} \hat{B} - I||_\infty \leq \lambda_n$. So by the definition of $\{\widehat{\Omega}_{tot}^1\}$, we have that

$$\left|\left|\widehat{\Omega}_{tot}^1\right|\right|_1 \leq ||\widehat{\mathbf{B}}||_1. \tag{31}$$

With Inequality (30) and Inequality (31), we have that $\widehat{\mathbf{B}} \in \{\widehat{\Omega}_{tot}^1\}$. On the other hand, if $\widehat{\Omega}_{tot}^1 \notin \{\widehat{\mathbf{B}}\}$, then $\exists i$ such that $||\hat{\omega}_j^1||_1 > ||\hat{\beta}_j + \hat{\beta}_j^s||_1$. Thus by Inequality (30), we have that $|\widehat{\Omega}^1|_1 > ||\widehat{\mathbf{B}}||_1$. This is in conflict with Inequality (31). □

## 7.2 Theorem 3

*Proof* By the condition in Theorem 3,

$$||\Sigma_{tot}^0 - \Sigma_{tot}||_\infty \leq \frac{\lambda_n}{||\Omega_{tot}^0||_1}. \tag{32}$$

we then have

$$\begin{aligned} \left|\left|I - \Sigma_{tot} \Omega_{tot}^0\right|\right|_\infty &= \left|\left|\left(\Sigma_{tot}^0 - \Sigma_{tot}\right) \Omega_{tot}^0\right|\right|_\infty \\ &\leq \left|\left|\Omega_{tot}^0\right|\right|_1 \left|\left|\Sigma_{tot}^0 - \Sigma_{tot}\right|\right|_\infty \leq \lambda_n. \end{aligned} \tag{33}$$

where $|\mathbf{AD}|_\infty \leq |\mathbf{A}|_\infty |\mathbf{D}|_1$ for any matrices $\mathbf{A}, \mathbf{D}$ of appropriate sizes.

Since $\Omega_{tot}^0$ also satisfies Eq. (16) and $\widehat{\Omega}_{tot}^1$ is the optimal solution which satisfies Eq. (16)

$$\left|\left|\widehat{\Omega}_{tot}^1\right|\right|_1 \leq \left|\left|\Omega_{tot}^0\right|\right|_1. \tag{34}$$

we have that

$$\begin{aligned} &\left|\left|\Sigma_{tot} \left(\widehat{\Omega}_{tot}^1 - \Omega_{tot}^0\right)\right|\right|_\infty \\ &\leq \left|\left|\Sigma_{tot} \widehat{\Omega}_{tot}^1 - I\right|\right|_\infty + \left|\left|I - \Sigma_{tot} \Omega_{tot}^0\right|\right|_\infty \leq 2\lambda_n. \end{aligned} \tag{35}$$

Therefore, based on Inequality (32) to Inequality (35)

$$||\Sigma_{tot}^0(\widehat{\Omega}_{tot}^1 - \Omega_{tot}^0)||_\infty \leq ||\Sigma_{tot}(\widehat{\Omega}_{tot}^1 - \Omega_{tot}^0)||_1 + ||(\Sigma_{tot} - \Sigma_{tot}^0)(\widehat{\Omega}_{tot}^1 - \Omega_{tot}^0)||_\infty$$
$$\leq 2\lambda_n + ||\widehat{\Omega}_{tot}^1 - \Omega_{tot}^0||_1 ||\Sigma_{tot} - \Sigma_{tot}^0||_\infty \leq 4\lambda_n.$$

Then,

$$||\widehat{\Omega}_{tot}^1 - \Omega_{tot}^0||_\infty \leq ||\Omega_{tot}^0||_1 ||\Sigma_{tot}^0(\widehat{\Omega}_{tot}^1 - \Omega_{tot}^0)||_\infty \leq 4 ||\Omega_{tot}^0||_1 \lambda_n.$$

This proves Inequality (20) of Theorem 3.

To prove the Inequality (21), let $t_n = ||\widehat{\Omega}_{tot} - \Omega_{tot}^0||_\infty$ and define

$$\mathbf{h}_j = \hat{\omega}_j - \omega_j^0$$
$$\mathbf{h}_j^1 = (\hat{\omega}_{ij} I\{|\hat{\omega}_{ij}| \geq 2t_n\}; 1 \leq i \leq p)^T - \omega_j^0$$
$$\mathbf{h}_j^2 = \mathbf{h}_j - \mathbf{h}_j^1.$$

By the definition of $\widehat{\Omega}_{tot}$, we have that $||\hat{\omega}_j||_1 \leq ||\hat{\omega}_j^0||_1 \leq ||\omega_j^0||_1$. Then

$$\left|\left|w_j^0\right|\right|_1 - \left|\left|\mathbf{h}_j^1\right|\right|_1 + \left|\left|\mathbf{h}_j^2\right|\right|_1 \leq \left|\left|\omega_j^0 + \mathbf{h}_j^1\right|\right|_1 + \left|\left|\mathbf{h}_j^2\right|\right|_1 = ||\hat{\omega}_{ij}||_1 \leq \left|\left|\omega_j^0\right|\right|_1,$$

which implies that $||\mathbf{h}_j^2||_1 \leq ||\mathbf{h}_j^1||_1$. It follows that $||\mathbf{h}_j||_1 \leq 2||\mathbf{h}_j^1||_1$. Thus we only need to get the upper bound of $||\mathbf{h}_j^1||_1$. We have that

$$\left|\left|\mathbf{h}_j^1\right|\right|_1 = \sum_{i=1}^p \left|\hat{\omega}_{ij} I\left\{|\hat{\omega}_{ij}| \geq 2t_n\right\} - \omega_{ij}^0\right|$$

$$\leq \sum_{i=1}^p \left|\omega_{ij}^0 I\left\{|\omega_{ij}^0| \leq 2t_n\right\}\right| + \sum_{i=1}^p \left|\hat{\omega}_{ij} I\left\{|\hat{\omega}_{ij}| \geq 2t_n\right\} - \omega_{ij}^0 I\left\{|\omega_{ij}^0| \leq 2t_n\right\}\right|$$

$$\leq (2t_n)^{1-q} s_0(p) + t_n \sum_{i=1}^p I\left\{|\hat{\omega}_{ij}| \geq 2t_n\right\}$$

$$+ \sum_{i=1}^p \left|\omega_{ij}^0\right| \left|I\left\{|\hat{\omega}_{ij}| \geq 2t_n\right\} - \omega_{ij}^0 - I\left\{|\omega_{ij}^0| \leq 2t_n\right\}\right|$$

$$\leq 2(t_n)^{1-q} s_0(p) + t_n \sum_{i=1}^p I\left\{|\omega_{ij}^0| \geq t_n\right\}$$

$$+ \sum_{i=1}^p \left|\omega_{ij}^0\right| I\left\{\left||w_{ij}^0| - 2t_n\right| \leq \left|\hat{\omega}_{ij} - \omega_{ij}^0\right|\right\}$$

$$\leq (2t_n)^{1-q} s_0(p) + (t_n)^{1-q} s_0(p) + (3t_n)^{1-q} s_0(p)$$

$$\leq (1 + 2^{1-q} + 3^{1-q}) t_n^{1-q} s_0(p). \qquad (36)$$

This proves the Inequality (21) of Theorem 3. For any $a, b, c \in \mathbb{R}$, we have that $|I\{a < c\} - I\{ab < c\}| \leq I\{|b - c| < |a - b|\}$

Finally, Inequality (22) can be derived from Inequality (20), Inequality (36) and the inequality relationship $||\mathbf{A}||_F^2 \leq p||\mathbf{A}||_1 ||\mathbf{A}||_\infty$. □

### 7.3 Theorem 4

*Proof* Theorem 4(a)

In the beginning of Sect. 5, we define that $\Sigma_{tot} := (\sigma_{ij})_{Kp \times Kp}$. By Theorem 3, we only need to prove

$$\max_{ij} |\hat{\sigma}_{ij} - \sigma_{ij}^0| \le C_0 \sqrt{\log Kp/n} \tag{37}$$

with probability greater than $1 - 4p^{-\tau_0}$ under (**C1**). Without loss of generality, we assume that $\mathbb{E}X^{(i)} = 0$. Let $\hat{\Sigma}_{tot}^0 := n_{tot}^{-1} \sum_{k=1}^{n} \mathbf{X}_{tot\,k} \mathbf{X}_{tot\,k}^T$ and $\mathbf{Y}_{kij} = \mathbf{X}_{tot\,ki} \mathbf{X}_{tot\,kj} - \mathbb{E}\mathbf{X}_{tot\,ki} \mathbf{X}_{tot\,kj}$. We then have $\hat{\Sigma}_{tot} = \hat{\Sigma}_{tot}^0 - \bar{\mathbf{X}}_{tot} \bar{\mathbf{X}}_{tot}^T$. Let $t = \eta \sqrt{\log Kp/n_{tot}}$.

Now we use the inequality $|e^s - 1 - s| < s^2 e^{\max(s,0)}$ for any $s \in \mathbb{R}$ and let

$$C_{K1} = 2 + \tau_0 + \eta^{-1}C^2. \tag{38}$$

with a few step of basic calculations we can get

$$\mathbb{P}\left(\sum_{k=1}^{n_{tot}} \mathbf{Y}_{kij} \ge \eta^{-1} C_{K1} \sqrt{n_{tot} \log Kp}\right)$$

$$\le e^{-C_{K1} \log Kp} (\mathbb{E}\exp(tY_{kij}))_{tot}^n$$

$$\le \exp\left(-C_{K1} \log Kp + n_{tot} t^2 \mathbb{E}Y_{kij}^2 e^{t|Y_{kij}|}\right)$$

$$\le \exp(-C_{K1} \log Kp + \eta^{-1}C^2 \log Kp)$$

$$\le \exp(-(\tau_0 + 2) \log Kp).$$

Thus, we have

$$\mathbb{P}\left(\left\|\hat{\Sigma}_{tot}^0 - \Sigma_{tot}^0\right\|_\infty \ge \eta^{-1} C_{K1} \sqrt{\log Kp/n_{tot}}\right) \le 2p^{-\tau_0}. \tag{39}$$

Then using the simple inequality $e^s \le e^{s^2+1}$ for $s > 0$, we have

$$\mathbb{E}e^{t|X_j|} \le eC, \quad \forall t \le \eta^{\frac{1}{2}}. \tag{40}$$

Let

$$C_{K2} = 2 + \tau_0 + \eta^{-1}e^2 C^2 \tag{41}$$

and $a_n = C_{K2}^2 (\log Kp/n_{tot})^{\frac{1}{2}}$. As before, we can show that

$$\mathbb{P}(\|\bar{\mathbf{X}}\bar{\mathbf{X}}^T\|_\infty \ge \eta^{-2} a_n \sqrt{\log Kp/n_{tot}})$$

$$\le p \max_i \mathbb{P}\left(\sum_{k=1}^{n_{tot}} \mathbf{X}_{ki} \ge \eta^{-1} C_{K2} \sqrt{n_{tot} \log Kp}\right)$$

$$+ p \max_i \mathbb{P}\left(-\sum_{k=1}^{n_{tot}} \mathbf{X}_{ki} \ge \eta^{-1} C_{K2} \sqrt{n_{tot} \log Kp}\right)$$

$$\le 2p^{-\tau_0-1}. \tag{42}$$

By Inequality (39), Inequality (42), and the inequality formulation $C_0 > \eta^{-1}C_{K1} + \eta^{-2}a_n$, we can prove that Inequality (37) holds. □

*Proof* Theorem [4](b)

Let $\breve{\mathbf{Y}}_{kij} = \mathbf{X}_{ki}\mathbf{X}_{kj}I\{|\mathbf{X}_{ki}\mathbf{X}_{kj}| \leq \sqrt{n_{tot}/(\log Kp)^3}\} - \mathbb{E}\mathbf{X}_{ki}\mathbf{X}_{kj}I\{|\mathbf{X}_{ki}\mathbf{X}_{kj}| \leq \sqrt{n_{tot}/(\log Kp)^3}\}$, $\hat{\mathbf{Y}}_{kij} = \mathbf{Y}_{kij} - \bar{\mathbf{Y}}_{kij}$.

Since

$b_n := \max_{i,j}\mathbb{E}|\mathbf{X}_{ki}\mathbf{X}_{kj}|I\{|\mathbf{X}_{ki}\mathbf{X}_{kj}| \leq \sqrt{n_{tot}/(\log Kp)^3}\} = O(1)n_{tot}^{-\gamma-\frac{1}{2}}$, we have, by condition (**C2**),

$$\mathbb{P}\left(\max_{i,j}\left|\sum_{k=1}^{n_{tot}}\breve{\mathbf{Y}}_{ijk}\right| \geq 2n_{tot}b_n\right)$$

$$\leq \mathbb{P}\left(\max_{i,j}\left|\sum_{k=1}^{n_{tot}}\mathbf{X}_{ki}\mathbf{X}_{kj}I\{|\mathbf{X}_{ki}\mathbf{X}_{kj}| > \sqrt{n_{tot}/(\log Kp)^3}\}|\right| \geq n_{tot}b_n\right)$$

$$\leq \mathbb{P}\left(\max_{i,j}\sum_{k=1}^{n_{tot}}|\mathbf{X}_{ki}\mathbf{X}_{kj}|I\{\mathbf{X}_{ki}^2 + \mathbf{X}_{kj}^2 \geq 2\sqrt{n_{tot}/(\log Kp)^3}\} \geq nb_n\right)$$

$$\leq \mathbb{P}\left(\max_{i,j}\mathbf{X}_{ki}^2 \geq \sqrt{n_{tot}/(\log Kp)^3}\right)$$

$$\leq pn_{tot}\mathbb{P}\left(\mathbf{X}_{11}^2 \geq \sqrt{n_{tot}/(\log Kp)^3}\right)$$

$$= O(1)n_{tot}^{-\frac{\delta}{8}}.$$

By Bernstein's inequality (cf. Bennett 1962) and some elementary calculations,

$$\mathbb{P}\left(\max_{i,j}\left|\sum_{k=1}^{n_{tot}}\bar{\mathbf{Y}}_{kij}\right| \geq \sqrt{(\theta_0+1)(4+\tau_0)n_{tot}\log Kp}\right)$$

$$\leq p^2\max_{i,j}\mathbb{P}\left(\left|\sum_{k=1}^{n_{tot}}\bar{\mathbf{Y}}_{kij}\right| \geq \sqrt{(\theta_0+1)(4+\tau_0)n_{tot}\log Kp}\right)$$

$$\leq 2p^2\max_{i,j}\exp(-(\theta_0+1)(4+\tau_0)n_{tot}\log Kp)/(2n_{tot}\mathbb{E}\bar{\mathbf{Y}}_{1ij}^2$$

$$+\sqrt{(\theta_0+1)(64+16\tau_0)}n_{tot}/(3\log Kp))$$

$$= O(1)p^{-\tau_0/2}.$$

Therefore, we have

$$\mathbb{P}\left(\left\|\hat{\Sigma}_{tot}^0 - \Sigma_{tot}^0\right\|_\infty \geq \sqrt{(\theta_0+1)(4+\tau_0)\log Kp/n_{tot}} + 2b_n\right)$$
$$= O\left(n_{tot}^{-\delta/8} + p^{-\tau_0/2}\right). \tag{43}$$

By using the same truncation argument and Bernstein's inequality, we can show that $\mathbb{P}(\max_i|\sum_{k=1}^{n_{tot}}\mathbf{X}_{ki}| \geq \sqrt{\max_i\sigma_{ii}^0(4+\tau_0)n_{tot}\log Kp}) = O(n_{tot}^{-\delta/8} + p^{-\tau_0/2})$.

Therefore,

$$\mathbb{P}\left(||\bar{\mathbf{X}}\bar{\mathbf{X}}^T||_\infty \geq \max_i\sigma_{ii}^0(4+\tau_0)\log Kp/n_{tot}\right) = O\left(n_{tot}^{-\delta/8} + p^{-\tau_0/2}\right) \tag{44}$$

Combining Eq. ([43]) and Eq. ([44]), then, we have

$$\max_{ij}|\hat{\sigma}_{ij} - \sigma_{ij}^0| \leq \sqrt{(\theta_0+1)(5+\tau_0)\log Kp/n_{tot}} \tag{45}$$

with a probability greater than $1 - O(n_{tot}^{-\delta/8} + p^{-\tau_0/2})$. The proof is completed by Inequality (45) and Theorem 3. □

### 7.4 Theorem 7

*Proof* Assume that there exists two different optimal solution $(\hat{\Omega}_I^{(i)}, \hat{\Omega}_S)$ and $(\hat{\Omega}_I^{(i)\prime}, \hat{\Omega}_S')$, which satisfies Eq. (7) and $\hat{\Omega}_I^{(i)} + \hat{\Omega}_S = \hat{\Omega}_I^{(i)\prime} + \hat{\Omega}_S'$. Since they are the optimal solutions, by Lemma 1, we have that $\hat{\Omega}_{I,j,k}^{(i)}\hat{\Omega}_{S,j,k} \geq 0$ and $\hat{\Omega}_{I,j,k}^{(i)\prime}\hat{\Omega}_{S,j,k}' \geq 0$. Therefore, $|\hat{\Omega}_{I,j,k}^{(i)}| + |\hat{\Omega}_{S,j,k}| = |\hat{\Omega}_{I,j,k}^{(i)} + \hat{\Omega}_{S,j,k}|$ and $|\hat{\Omega}_{I,j,k}^{(i)\prime}| + |\hat{\Omega}_{S,j,k}'| = |\hat{\Omega}_{I,j,k}^{(i)\prime} + \hat{\Omega}_{S,j,k}'|$. Now we get

$$||\Omega^{(i)}||_1 = \left|\left|\Omega_I^{(i)}\right|\right|_1 + ||\Omega_S||_1 \tag{46}$$

Then we denote $\hat{\Omega}^{(i)} = \hat{\Omega}_I^{(i)} + \hat{\Omega}_S$ and $\hat{\Omega}^{(i)\prime} = \hat{\Omega}_I^{(i)\prime} + \hat{\Omega}_S'$. Combining Eq. (7), Eq. (5) and Eq. (46), $(\hat{\Omega}^{(i)}, \hat{\Omega}_S)$ and $(\hat{\Omega}^{(i)\prime}, \hat{\Omega}_S')$ are both the optimal solution of

$$\underset{\Omega^{(i)}, \Omega_S}{\operatorname{argmin}} \sum_i ||\Omega^{(i)}||_1 + K(\varepsilon - 1)||\Omega_S||_1$$
$$\text{Subject to: } ||\Sigma^{(i)}\Omega^{(i)} - I||_\infty \leq \lambda_n, \quad i = 1, \dots, K \tag{47}$$

Because the constrains are not related to $\Omega_S$ and because we have that $\hat{\Omega}_{I,j,k}^{(i)} + \hat{\Omega}_{S,j,k} = \hat{\Omega}_{j,k}^{(i)} = \hat{\Omega}_{j,k}^{(i)\prime} = \hat{\Omega}_{I,j,k}^{(i)\prime} + \hat{\Omega}_{S,j,k}'$. By Lemma 1, $|\hat{\Omega}_{I,j,k}^{(i)}| + |\hat{\Omega}_{S,j,k}| = |\hat{\Omega}_{I,j,k}^{(i)\prime}| + |\hat{\Omega}_{S,j,k}'| = |\hat{\Omega}_{j,k}^{(i)}|$. When $\varepsilon < 1$, the optimization of Eq. (47) changes to the following:

$$\underset{\Omega_S}{\operatorname{argmax}} ||\Omega_S||_1$$
$$\text{Subject to: } \left|\Omega_{S,j,k}\right| \leq \left|\hat{\Omega}_{j,k}^{(i)}\right|, \Omega_{S,j,k}\hat{\Omega}_{j,k}^{(i)} \geq 0 \tag{48}$$

Equation (48) gives the unique solution $\hat{\Omega}_{S,j,k} = \min_i(|\hat{\Omega}_{j,k}^{(i)}|) \operatorname{sign}(\operatorname{argmin}_{\hat{\Omega}_{j,k}^{(i)}} |\hat{\Omega}_{j,k}^{(i)}|)$. The uniqueness contradicts the assumption of $\hat{\Omega}_S \neq \hat{\Omega}_S'$. For $\varepsilon > 1$, similarly we can prove, there exist unique $\Omega_S$ and $\{\Omega_I^{(i)}|i = 1, \dots, K\}$. □

## 8 Conclusions

This paper introduces a novel method SIMULE for learning shared and distinct patterns simultaneously when inferring multiple sGGMs or sNGMs jointly. Through $\ell 1$-constrained formulation our solution is efficient and can be parallelized. We successfully apply SIMULE on four synthetic datasets and two real-world datasets. We prove the convergence property of SIMULE to be favorable and justify the benefit of multitasking. Future work will extend SIMULE to model more complex relationships among contexts.

**Acknowledgements** This work was supported by the National Science Foundation under NSF CAREER Award No. 1453580. Any Opinions, findings and conclusions or recommendations expressed in this material are those of the author(s) and do not necessarily reflect those of the National Science Foundation.

# References

Antoniadis, A., & Fan, J. (2011). Regularization of wavelet approximations. *Journal of the American Statistical Association*, *96*(455), 939–967.

Banerjee, O., El Ghaoui, L., & d'Aspremont, A. (2008). Model selection through sparse maximum likelihood estimation for multivariate Gaussian or binary data. *The Journal of Machine Learning Research*, *9*, 485–516.

Boyd, S. P., & Vandenberghe, L. (2004). *Convex optimization*. Cambridge: Cambridge University Press.

Buchman, D., Schmidt, M., Mohamed, S., Poole, D., & de Freitas, N. (2012). On sparse, spectral and other parameterizations of binary probabilistic models. In *AISTATS* (pp. 173–181)

Cai, T., Liu, W., & Luo, X. (2011). A constrained 1 minimization approach to sparse precision matrix estimation. *Journal of the American Statistical Association*, *106*(494), 594–607.

Candes, E., & Tao, T. (2007). The Dantzig selector: Statistical estimation when p is much larger than n. *The Annals of Statistics*, *35*(6), 2313–2351.

Caruana, R. (1997). Multitask learning. *Machine Learning*, *28*(1), 41–75.

Cheng, C., Yan, K. K., Hwang, W., Qian, J., Bhardwaj, N., Rozowsky, J., et al. (2011). Construction and analysis of an integrated regulatory network derived from high-throughput sequencing data. *PLoS Computational Biology*, *7*(11), e1002190.

Chiquet, J., Grandvalet, Y., & Ambroise, C. (2011). Inferring multiple graphical structures. *Statistics and Computing*, *21*(4), 537–553.

Da Wei Huang, B. T. S., & Lempicki, R. A. (2008). Systematic and integrative analysis of large gene lists using DAVID bioinformatics resources. *Nature Protocols*, *4*(1), 44–57.

Danaher, P., Wang, P., & Witten, D. M. (2013). The joint graphical lasso for inverse covariance estimation across multiple classes. *Journal of the Royal Statistical Society: Series B (Statistical Methodology)*, *76*(2), 373–397.

Di Martino, A., Yan, C. G., Li, Q., Denio, E., Castellanos, F. X., Alaerts, K., et al. (2014). The autism brain imaging data exchange: Towards a large-scale evaluation of the intrinsic brain architecture in autism. *Molecular Psychiatry*, *19*(6), 659–667.

ENCODE Project Consortium. (2011). A user's guide to the encyclopedia of DNA elements (ENCODE). *PLoS Biology,9*(4), e1001046.

ENCODE Project Consortium. (2012). An integrated encyclopedia of DNA elements in the human genome. *Nature*, *489*(7414), 57–74.

Evgeniou, T., & Pontil, M. (2004). Regularized multi-task learning. In *Proceedings of the tenth ACM SIGKDD international conference on Knowledge discovery and data mining* (pp. 109–117). ACM.

Fan, J., Han, F., & Liu, H. (2014). Challenges of big data analysis. *National Science Review*,. doi:10.1093/nsr/nwt032.

Fan, J., Liao, Y., & Mincheva, M. (2013). Large covariance estimation by thresholding principal orthogonal complements. *Journal of the Royal Statistical Society: Series B (Statistical Methodology)*, *75*(4), 603–680.

Fazayeli, F., & Banerjee, A. (2016). Generalized direct change estimation in ising model structure. arXiv preprint arXiv:1606.05302.

Friedman, J., Hastie, T., & Tibshirani, R. (2008). Sparse inverse covariance estimation with the graphical lasso. *Biostatistics*, *9*(3), 432–441.

Guo, J., Levina, E., Michailidis, G., & Zhu, J. (2011). Joint estimation of multiple graphical models. *Biometrika*,. doi:10.1093/biomet/asq060.

Han, F., Liu, H., & Caffo, B. (2013). Sparse median graphs estimation in a high dimensional semiparametric model. arXiv preprint arXiv:1310.3223.

Hara, S., & Washio, T. (2013). Learning a common substructure of multiple graphical Gaussian models. *Neural Networks*, *38*, 23–38.

Hastie, T., Tibshirani, R., Friedman, J., Hastie, T., Friedman, J., & Tibshirani, R. (2009). *The elements of statistical learning*. Berlin: Springer.

Höfling, H., & Tibshirani, R. (2009). Estimation of sparse binary pairwise markov networks using pseudolikelihoods. *The Journal of Machine Learning Research*, *10*, 883–906.

Honorio, J., & Samaras, D. (2010). Multi-task learning of Gaussian graphical models. In *Proceedings of the 27th international conference on machine learning (ICML-10)* (pp. 447–454).

Hsieh, C. J., Sustik, M. A., Dhillon, I. S., & Ravikumar, P. D. (2011). Sparse inverse covariance matrix estimation using quadratic approximation. In *NIPS* (pp. 2330–2338).

Huang, S., Li, J., Sun, L., Ye, J., Fleisher, A., Wu, T., et al. (2010). Learning brain connectivity of alzheimer's disease by sparse inverse covariance estimation. *NeuroImage*, *50*(3), 935–949.

Ideker, T., & Krogan, N. J. (2012). Differential network biology. *Molecular Systems Biology*, *8*(1), 565.

Kelly, C., Biswal, B. B., Craddock, R. C., Castellanos, F. X., & Milham, M. P. (2012). Characterizing variation in the functional connectome: Promise and pitfalls. *Trends in Cognitive Sciences*, 16(3), 181–188.

Kolar, M., Song, L., Ahmed, A., Xing, E. P., et al. (2010). Estimating time-varying networks. *The Annals of Applied Statistics*, 4(1), 94–123.

Krumsiek, J., Suhre, K., Illig, T., Adamski, J., & Theis, F. J. (2011). Gaussian graphical modeling reconstructs pathway reactions from high-throughput metabolomics data. *BMC Systems Biology*, 5(1), 21.

Lam, C., & Fan, J. (2009). Sparsistency and rates of convergence in large covariance matrix estimation. *Annals of Statistics*, 37(6B), 4254.

Lauritzen, S. L. (1996). *Graphical models*. Oxford: Oxford University Press.

Levina, E., Rothman, A., Zhu, J., et al. (2008). Sparse estimation of large covariance matrices via a nested lasso penalty. *The Annals of Applied Statistics*, 2(1), 245–263.

Liu, H., Han, F., & Zhang, C. (2012). Transelliptical graphical models. In *Advances in Neural Information Processing Systems* (pp. 809–817).

Liu, H., Lafferty, J., & Wasserman, L. (2009). The nonparanormal: Semiparametric estimation of high dimensional undirected graphs. *The Journal of Machine Learning Research*, 10, 2295–2328.

Liu, H., Wang, L., & Zhao, T. (2014). Sparse covariance matrix estimation with eigenvalue constraints. *Journal of Computational and Graphical Statistics*, 23(2), 439–459.

Liu, S., Quinn, J. A., Gutmann, M. U., & Sugiyama, M. (2013). Direct learning of sparse changes in Markov networks by density ratio estimation. In *Joint European conference on machine learning and knowledge discovery in databases* (pp. 596–611). Springer.

Ma, S., Gong, Q., & Bohnert, H. J. (2007). An arabidopsis gene network based on the graphical Gaussian model. *Genome Research*, 17(11), 1614–1625.

Mardia, K. V., Kent, J. T., & Bibby, J. M. (1980). *Multivariate analysis*. London: Academic Press.

McCall, M. N., Uppal, K., Jaffee, H. A., Zilliox, M. J., & Irizarry, R. A. (2011). The gene expression barcode: Leveraging public data repositories to begin cataloging the human and murine transcriptomes. *Nucleic Acids Research*, 39(suppl 1), D1011–D1015.

Meinshausen, N., & Bühlmann, P. (2006). High-dimensional graphs and variable selection with the lasso. *The Annals of Statistics*, 34(3), 1436–1462.

Min, M. R., Ning, X., Cheng, C., & Gerstein, M. (2014). Interpretable sparse high-order Boltzmann machines. In *Proceedings of the seventeenth international conference on artificial intelligence and statistics* (pp. 614–622).

Mohan, K., London, P., Fazel, M., Lee, S. I., & Witten, D. (2013). Node-based learning of multiple Gaussian graphical models. arXiv preprint arXiv:1303.5145.

Monti, R. P., Anagnostopoulos, C., & Montana, G. (2015). Learning population and subject-specific brain connectivity networks via mixed neighborhood selection. arXiv preprint arXiv:1512.01947.

Negahban, S., Yu, B., Wainwright, M. J., & Ravikumar, P. K. (2009). A unified framework for high-dimensional analysis of *m*-estimators with decomposable regularizers. In *Advances in Neural Information Processing Systems* (pp. 1348–1356).

Ng, B., Varoquaux, G., Poline, J. B., & Thirion, B. (2013). A novel sparse group Gaussian graphical model for functional connectivity estimation. In *Information processing in medical imaging* (pp. 256–267). Springer.

Orchard, S., Ammari, M., Aranda, B., Breuza, L., Briganti, L., Broackes-Carter, F., et al. (2013). The MIntAct project IntAct as a common curation platform for 11 molecular interaction databases. *Nucleic Acids Research*, doi:10.1093/nar/gkt1115.

Pang, H., Liu, H., & Vanderbei, R. (2014). The fastclime package for linear programming and large-scale precision matrix estimation in R. *Journal of Machine Learning Research*, 15, 489–493.

Prasad, T. K., Goel, R., Kandasamy, K., Keerthikumar, S., Kumar, S., Mathivanan, S., et al. (2009). Human protein reference database 2009 update. *Nucleic Acids Research*, 37(suppl 1), D767–D772.

Qiu, H., Han, F., Liu, H., & Caffo, B. (2013). Joint estimation of multiple graphical models from high dimensional time series. arXiv preprint arXiv: 1311.0219.

Ripley, B. D. (2009). *Stochastic simulation* (Vol. 316). London: Wiley.

Rothman, A. J. (2012). Positive definite estimators of large covariance matrices. *Biometrika*, 99(3), 733–740.

Rothman, A. J., Bickel, P. J., Levina, E., Zhu, J., et al. (2008). Sparse permutation invariant covariance estimation. *Electronic Journal of Statistics*, 2, 494–515.

Schmidt, M., & Murphy, K. (2010). Convex structure learning in log-linear models: Beyond pairwise potentials. In *Proceedings of the international conference on artificial intelligence and statistics (AISTATS)*.

Stark, C., Breitkreutz, B. J., Reguly, T., Boucher, L., Breitkreutz, A., & Tyers, M. (2006). Biogrid: A general repository for interaction datasets. *Nucleic Acids Research*, 34(suppl 1), D535–D539.

Sugiyama, M., Kanamori, T., Suzuki, T., du Plessis, M. C., Liu, S., & Takeuchi, I. (2013). Density-difference estimation. *Neural Computation*, 25(10), 2734–2775.

Sun, L., Patel, R., Liu, J., Chen, K., Wu, T., Li, J., et al. (2009). Mining brain region connectivity for Alzheimer's disease study via sparse inverse covariance estimation. In *Proceedings of the 15th ACM SIGKDD international conference on Knowledge discovery and data mining* (pp. 1335–1344). ACM.

The Cancer Genome Atlas Research Network. (2011). Integrated genomic analyses of ovarian carcinoma. *Nature*, *474*(7353), 609–615.

Tibshirani, R. (1996). Regression shrinkage and selection via the lasso. *Journal of the Royal Statistical Society. Series B (Methodological)*, 267–288.

Wainwright, M. J., & Jordan, M. I. (2006). Log-determinant relaxation for approximate inference in discrete Markov random fields. *IEEE Transactions on Signal Processing*, *54*(6), 2099–2109.

Yang, E., Lozano, A. C., & Ravikumar, P. K. (2014). Elementary estimators for graphical models. In *Advances in neural information processing systems* (pp. 2159–2167).

Yuan, M., & Lin, Y. (2007). Model selection and estimation in the Gaussian graphical model. *Biometrika*, *94*(1), 19–35.

Zhang, B., & Wang, Y. (2012). Learning structural changes of Gaussian graphical models in controlled experiments. arXiv preprint arXiv:1203.3532.

Zhang, C. H. (2010). Nearly unbiased variable selection under minimax concave penalty. *The Annals of statistics*, *38*(2), 894–942.

Zhang, Y., & Schneider, J. G. (2010). Learning multiple tasks with a sparse matrix-normal penalty. In *Advances in neural information processing systems* (pp. 2550–2558).

Zhu, Y., Shen, X., & Pan, W. (2014). Structural pursuit over multiple undirected graphs. *Journal of the American Statistical Association*, *109*(508), 1683–1696.